\documentclass{article}

\usepackage{PRIMEarxiv}

\usepackage[utf8]{inputenc} 
\usepackage[T1]{fontenc}    
\usepackage{hyperref}       
\usepackage{url}            
\usepackage{booktabs}       
\usepackage{amsfonts}       
\usepackage{nicefrac}       
\usepackage{microtype}      
\usepackage{lipsum}
\usepackage{fancyhdr}       
\usepackage{graphicx}       
\graphicspath{{media/}}     

\usepackage{bm}
\usepackage{amssymb}
\usepackage{multirow}
\usepackage{amsmath}
\usepackage{pifont}
\newcommand{\cmark}{\ding{51}}
\newcommand{\xmark}{\ding{55}}

\title{UAV-Track VLA: Embodied Aerial Tracking via Vision-Language-Action Models}

\author{
  Qiyao Zhang$^{1, 7}$\thanks{The work was done during the internship at Institute of Automation, Chinese Academy of Sciences (3220241221\@bit.edu.cn).} \quad
  Shuhua Zheng$^{1}$ \quad
  Jianli Sun$^{2, 7}$ \quad
  Chengxiang Li$^{3, 7}$ \quad
  Xianke Wu$^{4, 7}$ \\
  \textbf{Zihan Song}$^{5, 7}$ \quad 
  \textbf{Zhiyong Cui}$^{6}$ \quad
  \textbf{Yisheng Lv}$^{2}$ \quad
  \textbf{Yonglin Tian}$^{2, 7}$\thanks{Corresponding author (yonglin.tian@ia.ac.cn).} \\
  \\ 
  $^{1}$Beijing Institute of Technology, Beijing, China \\
  $^{2}$Institute of Automation, Chinese Academy of Sciences, Beijing, China \\
  $^{3}$University of Sanya, Sanya, Hainan, China \\
  $^{4}$Beijing University of Posts and Telecommunications, Beijing, China \\
  $^{5}$Hunan University, Changsha, Hunan, China \\
  $^{6}$Beihang University, Beijing, China \\
$^{7}$Flying Intelligence Team (Virtual Research Community) \href{https://flying-intelligence.github.io/}{[url]}  \\
}
\begin{document}
\maketitle
\vspace{-1.0em}
\begin{figure*}[h]
    \centering
    \includegraphics[width=\textwidth]{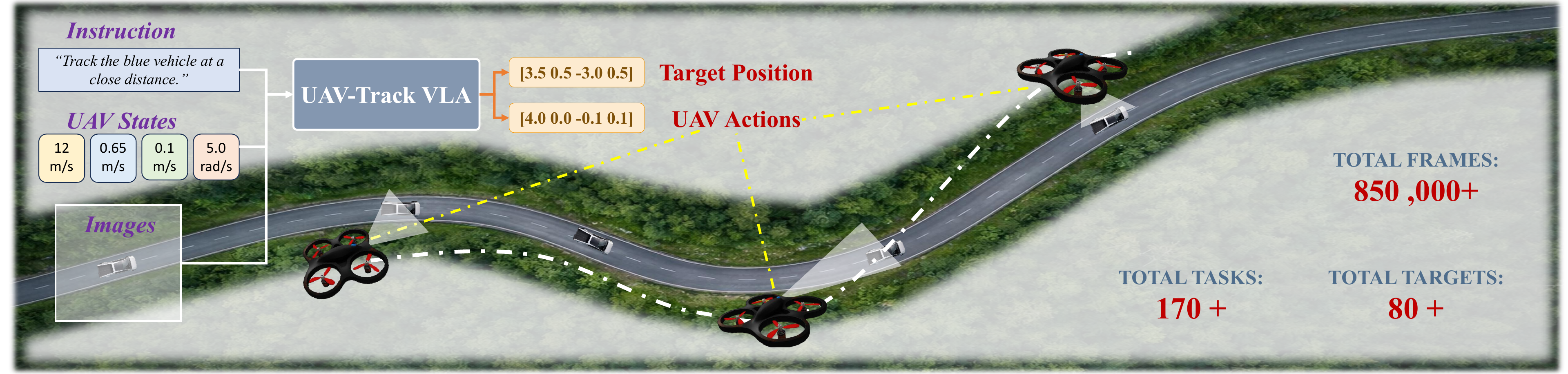}
    \caption{The UAV-Track VLA model follows human instructions to simultaneously predict the target position and continuous UAV flight actions.}
    \label{fig:dataset_diversity}
\end{figure*}

\begin{abstract}
Embodied visual tracking is crucial for Unmanned Aerial Vehicles (UAVs) executing complex real-world tasks. In dynamic urban scenarios with complex semantic requirements, Vision-Language-Action (VLA) models show great promise due to their cross-modal fusion and continuous action generation capabilities. To benchmark multimodal tracking in such environments, we construct a dedicated evaluation benchmark and a large-scale dataset encompassing over 890K frames, 176 tasks, and 85 diverse objects. Furthermore, to address temporal feature redundancy and the lack of spatial geometric priors in existing VLA models, we propose an improved VLA tracking model, UAV-Track VLA. Built upon the $\pi_{0.5}$ architecture, our model introduces a temporal compression net to efficiently capture inter-frame dynamics. Additionally, a parallel dual-branch decoder comprising a spatial-aware auxiliary grounding head and a flow matching action expert is designed to decouple cross-modal features and generate fine-grained continuous actions. Systematic experiments in the CARLA simulator validate the superior end-to-end performance of our method. Notably, in challenging long-distance pedestrian tracking tasks, UAV-Track VLA achieves a 61.76\% success rate and 269.65 average tracking frames, significantly outperforming existing baselines. Furthermore, it demonstrates robust zero-shot generalization in unseen environments and reduces single-step inference latency by 33.4\% (to 0.0571s) compared to the original $\pi_{0.5}$, enabling highly efficient, real-time UAV control. Data samples and demonstration videos are available at: https://github.com/Hub-Tian/UAV-Track\_VLA.
\end{abstract}

\keywords{Unmanned Aerial Vehicle \and Vision-Language-Action Models \and Embodied Visual Tracking}

\section{Introduction}
\maketitle

Unmanned Aerial Vehicles (UAVs) are being increasingly utilized to execute complex tasks in real-world environments \cite{tian2025uavs, 10445025}. As a core perceptual capability, visual tracking serves as the foundation for advanced applications such as inspection, search and rescue, and traffic monitoring\cite{khawaja2025survey}. However, contemporary UAV visual tracking predominantly relies on manual flight control, deploying deep learning models on the backend merely for passive target detection or segmentation on transmitted images \cite{Li_2020_CVPR, li2023adaptive, xue2024handling}. This decoupled paradigm fails to achieve autonomous pursuit and dynamic tracking, severely constraining the large-scale deployment and execution efficiency of UAVs in complex scenarios.

Driven by advancements in artificial intelligence, recent years have witnessed a surge in research on Visual Active Tracking (VAT) for UAVs \cite{9521193, sun2025openworld, dionigi2024d}. These approaches empower UAVs to autonomously make decisions and adjust flight postures based on visual feedback, thereby alleviating the reliance on manual control during tracking missions. Nevertheless, the vast majority of existing VAT frameworks are confined to a direct mapping between visual inputs and motor actions, completely omitting the language modality. This critical lack of high-level semantic understanding renders UAVs incapable of executing complex instructions specified by natural language, which in turn limits their applicability in real-world interactive scenarios.

With the rapid evolution of Large Language Models (LLMs) and multimodal fusion technologies, Vision-Language-Action (VLA) models have demonstrated remarkable capabilities in robotic manipulation, autonomous driving, and path planning \cite{brohan2023rt1roboticstransformerrealworld, kim2024openvlaopensourcevisionlanguageactionmodel, black2026pi0visionlanguageactionflowmodel,jiang2025survey, liu2025indooruavbenchmarkingvisionlanguageuav}. In the realm of object tracking, researchers have recently introduced the concept of Embodied Visual Tracking (EVT), formally integrating the language modality into active visual tracking. For instance, pioneering works like TrackVLA\cite{pmlr-v305-wang25f} have successfully incorporated VLA models into EVT tasks. However, current EVT research predominantly relies on ground-based platforms, such as quadrupedal or wheeled robots. Compared to ground robots constrained to a 2D plane, UAVs possess superior maneuverability in 3D space, offering a much freer tracking perspective. Despite this potential, there remains a glaring absence of dedicated EVT benchmarks and tailored methodologies for UAVs.

Urban outdoor scenarios are characterized by large spatial scales and pronounced target dynamics. Objects such as vehicles and pedestrians exhibit highly heterogeneous motion patterns, which imposes high requirements on the joint extraction of spatial and temporal information for continuous target tracking. Existing UAV active tracking datasets (such as those used in VLA-AN\cite{wu2025vlaanefficientonboardvisionlanguageaction}) fail to include targets with complex kinematic behaviors like vehicles, making them ill-suited for the demands of urban traffic scenarios. To train a versatile model capable of robust tracking across diverse targets, it is imperative to construct an EVT dataset that encompasses precise linguistic instructions, heterogeneous tracking objects, and supports continuous UAV motion control.

Furthermore, existing VLA models suffer from significant limitations in dynamic tracking tasks: they not only lack the capacity to process continuous temporal image sequences due to an over-reliance on single-frame spatial feature fusion, but also exhibit poor feature extraction capabilities for high-speed and irregularly moving objects. Consequently, these models struggle to align high-level semantics with low-level continuous flight control, making architectural improvements urgently needed.

To bridge these gaps, we propose a novel embodied UAV tracking benchmark, dubbed UAV-Track, tailored for complex urban scenarios, and introduce structural enhancements to existing VLA models to rectify their shortcomings in temporal feature extraction and cross-modal control alignment. The main contributions of this paper are summarized as follows:

\begin{itemize}
  \item We construct the first vision-language dataset for embodied UAV tracking in urban environments. Collected within the CARLA simulator, this dataset comprises over 890,000 frames, fine-grained spatial instructions, and diverse tracking targets including vehicles and pedestrians, thereby filling the void in semantic-level EVT benchmarks for UAVs.
  \item We propose a novel uav tracking vla method with temporal enhancement and spatial guidance. This model incorporates a temporal compression network to extract historical motion evolution patterns. Concurrently, it employs a parallel dual-branch decoding architecture: a spatial-aware auxiliary grounding head injects geometric priors, while a flow matching action expert generates a 25-step continuous displacement sequence, effectively aligning visual semantics with UAV kinematic control.
  \item We conduct comprehensive experimental evaluations in CARLA. The results demonstrate our model's superior performance in cross-modal alignment, continuous tracking capability, and robust zero-shot generalization across varying environments, target speeds, and spatial instructions.
\end{itemize}

\begin{figure*}[htbp]
    \centering
    \includegraphics[width=\textwidth]{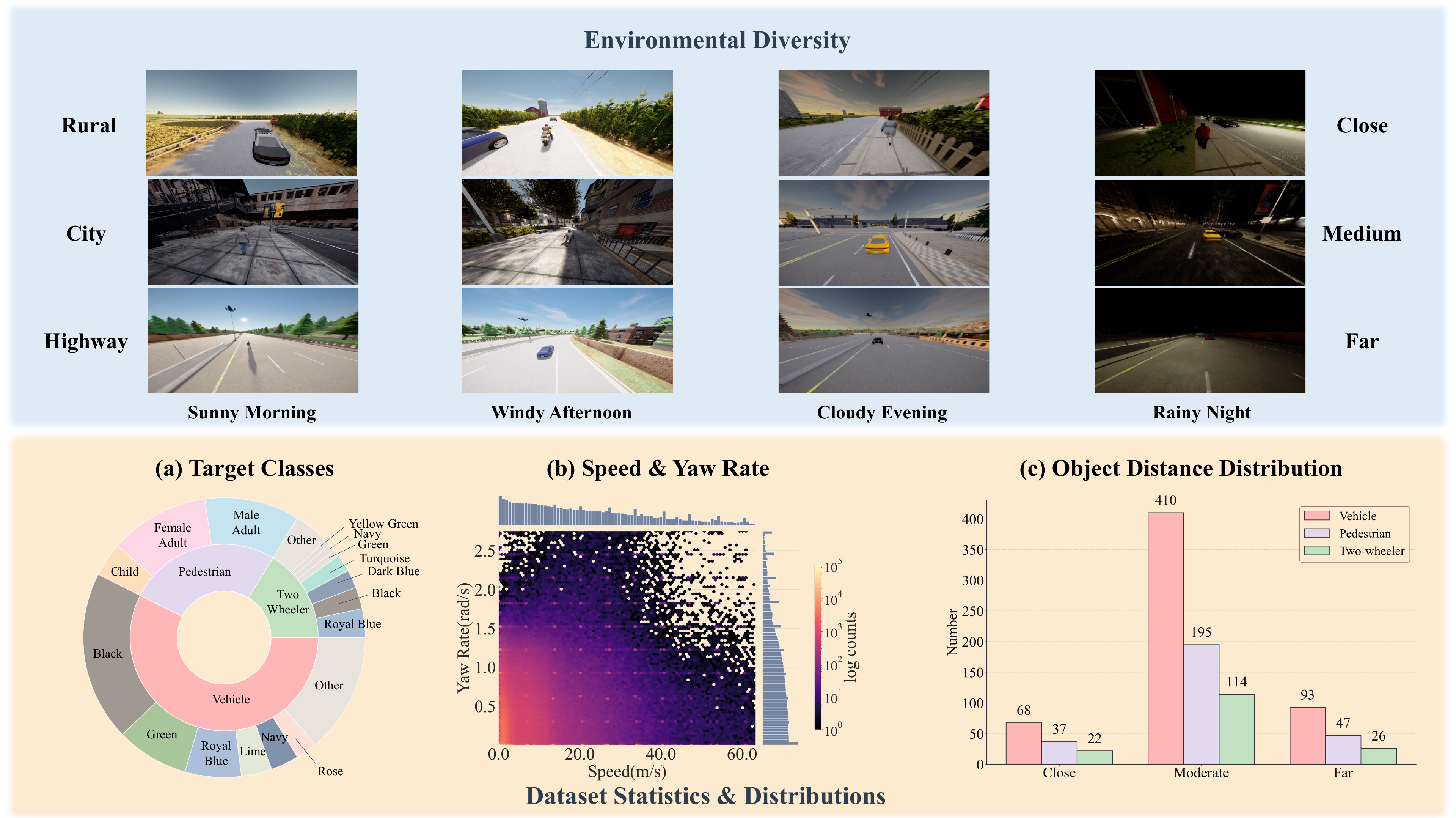}
    \caption{A comprehensive overview of the multidimensional diversity of our benchmark dataset. Top: Visualization of environment diversity. Middle: Task paradigms and overall dataset scale of the UAV-Track benchmark. Bottom: Quantitative distribution of target categories, kinematic parameters, and tracking distances.}
    \label{fig:dataset_diversity}
\end{figure*}

\section{Related Work}

\subsection{Embodied Visual Tracking}

As a crucial branch of computer vision, UAV Visual Tracking has advanced significantly to address aerial visual characteristics and hardware constraints \cite{Li_2020_CVPR,xue2024handling,li2023adaptive}. Recent research has shifted from merely improving tracking robustness \cite{sun2023uav,Li_2020_CVPR,xue2024handling} to optimizing model efficiency and semantic alignment \cite{wu2025learning,li2023adaptive,sun2025refdrone,Xue_2025_CVPR}. For instance, Aba-ViTrack \cite{li2023adaptive} reduces computational overhead via background aware token computation, while AVTrack-MD \cite{wu2025learning} employs multi-teacher knowledge distillation for efficient representation. Similarly, SGLATrack \cite{Xue_2025_CVPR} achieves an optimal accuracy-speed trade-off via dynamic layer pruning, and NGDINO \cite{sun2025refdrone} aligns visual tracking with natural language by explicitly modeling object counts. However, these methods remain within the traditional detection paradigm \cite{sun2023uav,Li_2020_CVPR,xue2024handling,wu2025learning,li2023adaptive,sun2025refdrone,Xue_2025_CVPR}, neglecting the UAV's intrinsic motion characteristics and the potential for embodied active tracking.

VAT, the predecessor to EVT, achieves continuous target locking through autonomous motion and perspective adjustments. Recent advancements in VAT encompass open-world benchmarks \cite{sun2025openworld}, 3D anti-distractor tracking \cite{9521193}, and end-to-end control \cite{8642452,dionigi2024d,Wu2025}. Specifically, GC-VAT \cite{sun2025openworld} utilizes goal-centered rewards and curriculum learning for dynamic environments, whereas Ad-AOT \cite{9521193} introduces reinforcement learning to navigate 3D distractors. D-VAT \cite{dionigi2024d} enables zero-fine-tuning simulation-to-real transfer for continuous micro-aerial vehicle control, and CEL \cite{Wu2025} resolves training divergence caused by anomalies via a cognitive embodied learning approach. Nevertheless, these VAT methods lack the capacity for complex natural language understanding, hindering their application in advanced semantic tasks.

EVT integrates perception and action decision-making to achieve continuous tracking through an embodied loop \cite{zhang2025uninavidvideobasedvisionlanguageactionmodel,wu2025hierarchicalinstructionawareembodiedvisual,11246600,liu2025trackvlaunleashingreasoningmemory,pmlr-v305-wang25f,Zhong_2025_ICCV}. To overcome VAT's semantic limitations, recent EVT research increasingly incorporates VLA models for instruction-driven tracking. For example, HIEVT \cite{wu2025hierarchicalinstructionawareembodiedvisual} uses spatial targets to bridge language and actions, while a VLM-enhanced framework \cite{11246600} introduces self-reflection to recover from tracking failures. Furthermore, TrackVLA \cite{pmlr-v305-wang25f} leverages a shared LLM backbone to synchronize target recognition with trajectory planning, while TrackVLA++ \cite{liu2025trackvlaunleashingreasoningmemory} introduces Polar Chain-of-Thought to enhance robustness against severe occlusions.

\subsection{VLA Models in Aerial Robotics}

The rapid progress in EVT is driven by VLA models' cross-modal fusion and continuous action generation capabilities \cite{cheang2025gr3technicalreport,kim2024openvlaopensourcevisionlanguageactionmodel,black2026pi0visionlanguageactionflowmodel}. Following the foundational multimodal synergy established by RT-1 and RT-2 \cite{brohan2023rt1roboticstransformerrealworld,brohan2023rt2visionlanguageactionmodelstransfer}, models like OpenVLA and $\pi_0$ utilized technologies such as action chunking, flow-matching, dual systems and diffusion architectures to achieve efficient continuous action generation under multimodal inputs \cite{octomodelteam2024octoopensourcegeneralistrobot,kim2024openvlaopensourcevisionlanguageactionmodel,black2026pi0visionlanguageactionflowmodel,song2025hume, han2024dualprocessvlaefficient,li2024cogactfoundationalvisionlanguageactionmodel}. Recently, the integration of reinforcement learning (e.g., $\pi_{0.6}$, GR-RL) has further propelled VLA models toward high-precision real-world manipulation \cite{cheang2025gr3technicalreport,intelligence2025pi05visionlanguageactionmodelopenworld,intelligence2025pi06vlalearnsexperience,li2025grrlgoingdexterousprecise}.

Leveraging their semantic and action-mapping strengths, VLA models are increasingly applied to UAVs. While current applications primarily focus on spatial navigation and obstacle avoidance \cite{liu2025indooruavbenchmarkingvisionlanguageuav,wang2025uavflowcolosseorealworldbenchmark,sun2026autoflyvisionlanguageactionmodeluav} using enhanced visual reasoning or hierarchical architectures \cite{wu2025vlaanefficientonboardvisionlanguageaction,sun2026autoflyvisionlanguageactionmodeluav,huang2026navdreamervideomodelszeroshot}, some studies have extended VLA to high-speed drone racing \cite{serpiva2025racevlavlabasedracingdrone}, real-time human recognition \cite{lykov2025cognitivedronevlamodelevaluation}, and aerial manipulation \cite{sun2026airvlavisionlanguageactionsystemsaerial}. However, existing end-to-end flight control methods mostly target static interactions and struggle with highly dynamic tracking. Although preliminary trajectory tracking exists \cite{wu2025vlaanefficientonboardvisionlanguageaction}, a substantial gap remains for urban scenarios, where vehicles and pedestrians exhibit severe motion randomness, frequent occlusions, and background interference. Developing an end-to-end VLA system capable of maintaining advanced semantic locking and ultra-low latency dynamic interaction in such volatile environments remains an urgent challenge that this paper seeks to address.

\section{UAV-Track Benchmark}
\begin{table*}[t]
\centering
\caption{Comparison of Existing UAV Tracking and Embodied AI Benchmarks}
\label{tab:benchmark_comparison}
\resizebox{\textwidth}{!}{%
\begin{tabular}{lccccc}
\toprule
\textbf{Benchmark} & \textbf{Emboied Tracking} & \textbf{Kinematic Diversity} & \textbf{Language} & \textbf{Continuous Motion} & \textbf{UAV Dedicated} \\ \midrule
DAT\cite{sun2025openworld} & \cmark & \cmark & \xmark & \cmark & \cmark \\
AD-AOT\cite{9521193} & \cmark & \xmark & \xmark & \xmark & \cmark \\
RefDrone\cite{sun2025refdrone} & \xmark & \cmark & \cmark & \xmark & \cmark \\
IndoorUAV\cite{liu2025indooruavbenchmarkingvisionlanguageuav} & \xmark & \xmark & \cmark & \cmark & \cmark \\
UAV-Flow/UAV-Flow-Sim\cite{wang2025uavflowcolosseorealworldbenchmark} & \xmark & \xmark & \cmark & \cmark & \cmark \\
VLA-AN\cite{wu2025vlaanefficientonboardvisionlanguageaction} & \cmark & \xmark & \cmark & \cmark & \cmark \\
EVT-Bench\cite{pmlr-v305-wang25f} & \cmark & \xmark & \cmark & \cmark & \xmark \\
UC-EVT\cite{wu2025hierarchicalinstructionawareembodiedvisual} & \cmark & \xmark & \cmark & \cmark & \xmark \\ 
\textbf{UAV-Track (Ours)} & \textbf{\cmark} & \textbf{\cmark} & \textbf{\cmark} & \textbf{\cmark} & \textbf{\cmark} \\ \bottomrule
\end{tabular}
}
\end{table*}

\subsection{Problem Formulation}

In this section, we introduce UAV-Track, a novel benchmark for natural language-guided Unmanned Aerial Vehicle visual tracking, which we formulate as an end-to-end sequential decision-making problem. By fusing the multimodal context $\mathcal{O}_t$, the VLA model directly outputs the relative pose of the target and the low-level action trajectory. At time step $t$, the system receives three types of input observations:

Language Instruction: $L = \{w_1, \dots, w_m\}$: Describes the target's appearance and the specific flight intention.
Visual Observations: $I_{t-3:t} \in \mathbb{R}^{4 \times H \times W \times 3}$: Comprises the current and three previous front-view RGB image frames (detailed parameters are provided in the supplementary material). For the initial stage where $t<4$, we employ a zero-padding strategy using black frames to maintain structural consistency in the temporal dimension.
Proprioceptive State: $S_t = [\bm{v}_t, \dot{\psi}_t]^\top \in \mathbb{R}^4$: Includes the 3D linear velocity (m/s) and the body yaw rate(rad/s). To reduce the complexity of policy learning, we fix the pitch and roll angles of the UAV as well as the camera extrinsics in the simulation.

Based on the above inputs, the model simultaneously performs spatial perception and flight control prediction:

Target Relative Pose Perception: Outputs the pose of the tracked target in the UAV-centric coordinate system, which is formulated as $P_t = [\Delta x^{\text{tar}}_t, \Delta y^{\text{tar}}_t, \Delta z^{\text{tar}}_t, \Delta \psi^{\text{tar}}_t]^\top \in \mathbb{R}^4$. This includes the 3D relative position and the horizontal yaw offset, providing precise geometric guidance.
Future Action Trajectory: Adopts an action chunking strategy to predict a sequence of continuous commands for the future $k$ steps ($k=25$ in our implementation), defined as $A_{t+1:t+k} = \{\bm{a}_t+1, \dots, \bm{a}_{t+k}\}$. The action vector $\bm{a}_i = [\Delta x^{\text{act}}_i, \Delta y^{\text{act}}_i, \Delta z^{\text{act}}_i, \Delta \psi^{\text{act}}_i]^\top \in \mathbb{R}^4$ directly represents the desired 3D relative displacement and yaw angle change, discarding traditional velocity control to perfectly align the action space with the perception output.

\subsection{Simulation and Data Collection}

To bridge the sim-to-real gap, we build a comprehensive data collection environment and processing pipeline based on CARLA that encompasses diverse topological scenes, dynamic weather, and varying lighting conditions. We adopt a hybrid strategy to collect large-scale trajectory data. First, human experts control the UAV to collect high-quality demonstrations, providing a baseline for spatial reasoning and fallback strategies for corner cases. Subsequently, we deploy an automated algorithm based on the Artificial Potential Field (APF) method for data augmentation\cite{jayaweera2020dynamic}. During automatic cruising, we dynamically inject random perturbations into the control commands to force the UAV off its ideal trajectory, and the APF generates recovery paths. This mechanism effectively alleviates the covariate shift problem common in end-to-end imitation learning, teaching the model to recover autonomously and significantly enhancing closed-loop robustness.

\subsection{Statistics}

This benchmark collects a total of 892,756 frames of multimodal trajectory data, comprising approximately 200K frames of expert demonstrations and 690K frames of automated collection data, covering 85 diverse objects and 176 fine-grained tracking tasks. As shown in Figure \ref{fig:dataset_diversity}, our dataset achieves comprehensive scenario coverage across three core dimensions:

Environment and Target Diversity: Covers dynamic weather, a full range of tracking distances, and three main categories of dynamic targets.
Kinematic Completeness: Encompasses a wide range of target speeds (0 - 70 m/s) and UAV yaw dynamics, supporting the learning of physically feasible policies.
Language Instruction Generalization: Hundreds of natural language instructions support cross-modal task evaluation.

Compared to existing benchmarks (Table \ref{tab:benchmark_comparison}), this work is the first dedicated emboied UAV tracking benchmark that comprehensively supports natural language guidance and 4-DoF continuous motion, demonstrating significant advantages in scene generalization and task universality.

\subsection{Evaluation Metrics}

We introduce two core metrics to measure the closed-loop tracking robustness of the system. Considering the complex occlusions in real-world scenarios, we allow the model to autonomously recover after briefly losing the target:

\subsubsection{Success Rate (SR)}

Given the total number of test episodes $N$, we define a binary indicator function $\mathbb{I}_{\text{success}}^{(i)}$. Episode $i$ is considered successful ($\mathbb{I}_{\text{success}}^{(i)} = 1$) only if it does not trigger a "fatal failure" during its execution period $T_i$. A fatal failure is defined as: the target continuously deviates from the effective distance range ($d_t \notin [d_{\min}, d_{\max}]$ or completely leaves the field of view) for a duration exceeding the maximum tolerance threshold $\tau$. The overall success rate is defined as $\text{SR} = \frac{1}{N} \sum_{i=1}^{N} \mathbb{I}_{\text{success}}^{(i)}$.

\subsubsection{Average Tracked Frames (ATF)}

ATF provides a more fine-grained process evaluation, measuring the average duration of maintaining effective tracking before a fatal failure occurs. For an episode with a maximum step limit of $T_i$, if a valid control is output and the target is within the effective range at a single step $t$, then $\mathbb{I}_{\text{track}}^{(i, t)} = 1$; once a fatal failure is triggered, all subsequent steps are set to zero. The metric is calculated as $\text{ATF} = \frac{1}{N} \sum_{i=1}^{N} \sum_{t=1}^{T_i} \mathbb{I}_{\text{track}}^{(i, t)}$. This metric intuitively reflects the differences in control stability among models when handling complex dynamics.

\begin{figure*}[htbp]
    \centering
    \includegraphics[width=0.95\textwidth]{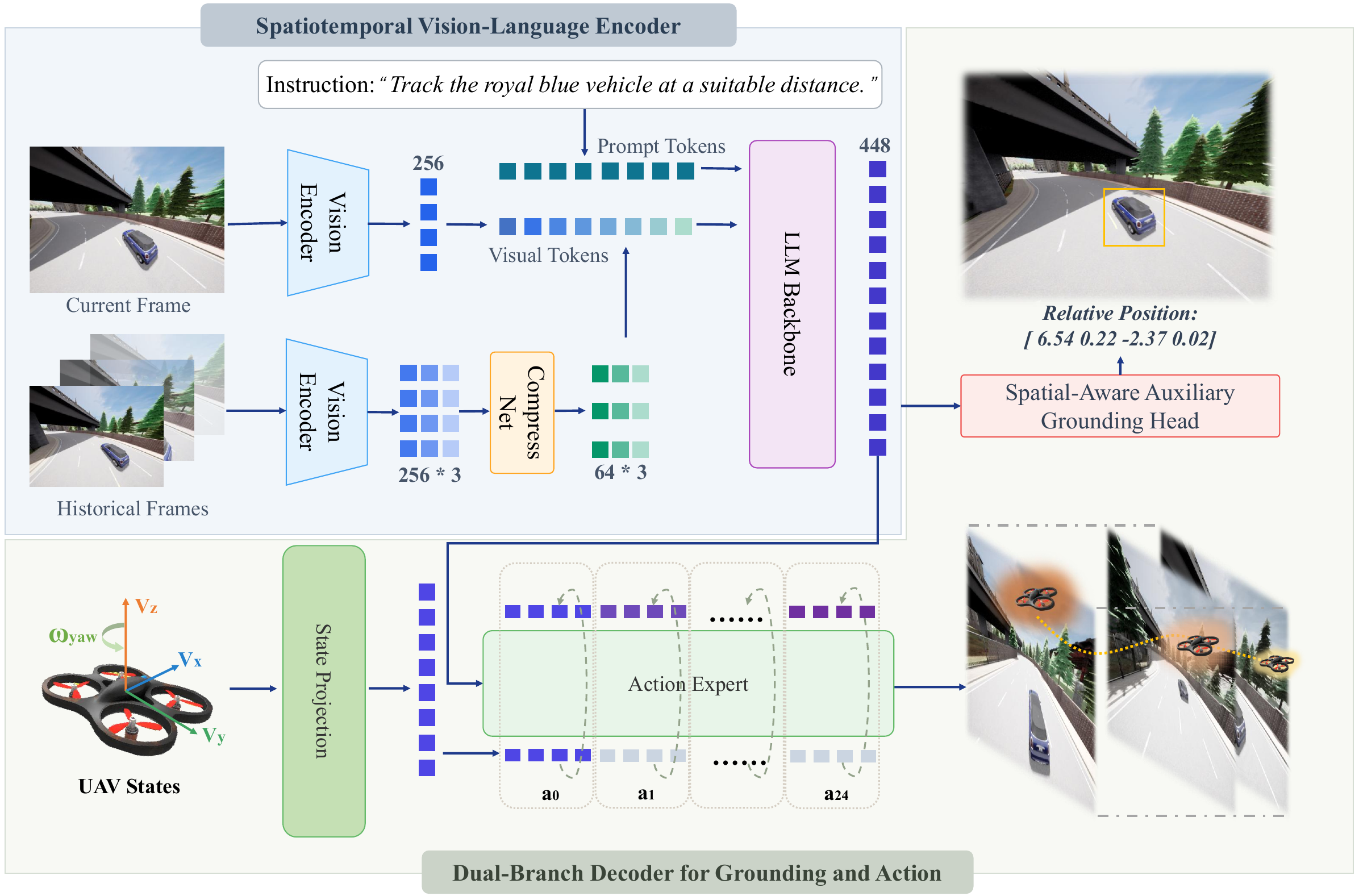}
    \caption{\textbf{Overall architecture of our proposed UAV-Track VLA model.} The model processes heterogeneous inputs through a spatiotemporal encoder and parallel decoders. Within the encoder, historical frames are temporally compressed and concatenated with the current frame to capture dynamic evolution, then fused with the instruction via an LLM. The shared multimodal features are subsequently dispatched to two decoupled branches: a spatial-aware auxiliary grounding head that predicts the target's relative pose $P_t$ to inject precise geometric priors for robust high-frequency control}
    \label{fig:method}
\end{figure*}

\section{UAV-Track VLA}

To address the limitations of general VLA models in continuous UAV tracking—specifically, their inadequate temporal modeling for historical frames and the severe mismatch between high-level semantic features and low-level continuous displacement control—we propose an improved UAV VLA tracking architecture tailored for complex urban scenarios. Built upon the open-source $\pi_{0.5}$ model, our approach introduces a temporal compression net and a spatial-aware auxiliary grounding head. By employing an end-to-end training strategy with a mixed loss function, our model effectively bridges the gap between high-level semantic understanding and high-frequency continuous control.

Our model adopts an encoder-dual-branch decoder hierarchical architecture. This design aims to assist the encoder in achieving precise spatial alignment of multimodal features without compromising the continuous control capabilities of the action expert. As illustrated in Figure \ref{fig:method}, the model takes heterogeneous inputs: a visual sequence containing current and historical observations $I_{t-3:t}$, a natural language instruction $L$, and the current proprioceptive state of the UAV $S_{t}$. After the encoder transforms these inputs into high-dimensional cross-modal features, two parallel decoding branches process them to output the target's relative pose $P_{t}$ for auxiliary supervision and a 25-step continuous action displacement sequence $A_{t+1:t+25}$ for high-frequency control.

\subsection{Spatiotemporal vision-language encoder}

The encoder integrates a visual feature extraction module, a temporal compression net, and a foundational large vision-language model (PaliGemma\cite{beyer2024paligemma}) to transform the heterogeneous inputs into high-dimensional cross-modal features enriched with temporal context and language semantics.

\subsubsection{Backbone}
Our model utilizes PaliGemma as its core foundational architecture. Specifically, the vision tower (SigLIP\cite{zhai2023sigmoid}) extracts image features, mapping single-frame images into visual tokens within the semantic feature space. Concurrently, the language instruction is processed by a tokenizer to generate instruction tokens. Subsequently, the Large Language Model (Gemma\cite{team2024gemma}) performs deep fusion and cross-modal interaction between the mapped visual semantic features and the textual semantic features.

\subsubsection{Temporal compression net}
Capturing the dynamic motion trends of targets necessitates the incorporation of historical visual information. To prevent the exponential growth in computational dimensions and feature redundancy caused by stacking multiple frames, we introduce a lightweight Temporal compression net prior to feature injection into Gemma. Specifically, for the three historical frames $I_{t-3:t-1}$, this module employs a linear projection layer to drastically compress the 256 visual tokens of each single frame down to 64 tokens. These $64 \times 3$ low-dimensional tokens are then concatenated with the 256 original tokens of the current frame along the temporal dimension, resulting in a cohesive set of 448 visual features embedded with temporal information. To enhance the model's ability to capture temporal dependencies across the input image sequence, we explicitly introduce a Learnable Positional Encoding, enabling the model to accurately extract the evolutionary patterns between frames. Ultimately, this comprehensive visual feature set is fed into Gemma alongside the instruction tokens for joint encoding.

\subsection{Dual-Branch decoder for grounding and action}

To prevent representational conflicts between explicit spatial grounding features and continuous action generation, we design a parallel dual-branch decoding structure comprising a spatial-aware auxiliary grounding head and a flow matching action expert. These two branches share the cross-modal features but remain completely decoupled during forward inference.

\subsubsection{Spatial-aware auxiliary grounding head}
To address the specific requirements of continuous tracking and the lack of precise spatial geometric priors when applying general VLA models to this task, we attach a compact transformer after Gemma to serve as the spatial-aware auxiliary grounding head. Taking cross-modal features as input, this module predicts the 3D relative position and horizontal yaw angle deviations of the target relative to the UAV. Crucially, this branch adopts a Bypass Auxiliary Design that operates exclusively via auxiliary supervision during the training phase. By backpropagating this spatial loss, the mechanism compels the encoder to embed accurate task-specific geometric priors into the cross-modal tokens, thereby establishing a robust feature foundation for subsequent continuous action planning.

\subsubsection{Flow matching action expert}
The action planning branch adopts a flow matching architecture, which effectively transforms discrete step predictions into flow field fitting within a continuous state space, highly aligning with the high-frequency continuous control demands of UAVs. This branch concatenates the cross-modal features—now imbued with spatial priors—with the UAV's proprioceptive state $S_{t}$. By learning to approximate the displacement flow field of the ideal tracking trajectory, the model performs continuous inference, ultimately generating a 25-step continuous displacement control sequence $A_{t+1:t+25}$.

\subsection{Objectives}

To fulfill the cross-modal alignment requirements of vision, language, and action in complex urban scenarios, we design an end-to-end joint training strategy based on a mixed loss function. This strategy synchronizes the optimization of the encoder, the spatial-aware auxiliary grounding head, and the flow matching action expert, achieving simultaneous cross-modal representation alignment and rational action generation within a single training process.

The total loss function of our model is a weighted fusion of the position loss and the flow matching loss. The core objective of this design is to optimize the effective mapping from spatial features to continuous control actions within the action expert, while leveraging the position loss to ensure the encoder's cross-modal features possess accurate geometric priors as a prerequisite.

The total loss function is defined as:
$$L_{total} = \lambda_{1} L_{pos} + \lambda_{2} L_{action}$$

where $L_{pos}$ denotes the position loss, which supervises the 3D position and yaw angle deviation predictions of the spatial-aware auxiliary grounding head, implicitly constraining the encoder's capacity to model spatial geometric information. $L_{action}$ represents the flow matching loss, acting as the core of the network optimization to supervise the generation quality of the continuous displacement sequence by the flow matching action expert, ensuring the physical rationality and smoothness of the UAV's 25-step action output. $\lambda_{1}$ and $\lambda_{2}$ are weight coefficients. By judiciously adjusting their ratio, the model prioritizes enhancing the generation quality of the action expert's continuous control, while utilizing the spatial grounding supervision to provide a reliable feature foundation, ultimately achieving optimal overall performance in multimodal continuous tracking.

\section{Experiments}

\subsection{Experimental Setup}

To systematically evaluate the performance of the proposed UAV-Track VLA in complex dynamic environments, all experiments are conducted within the CARLA simulator. At the onset of each evaluation episode, the UAV is initialized at a random position behind and above the target to perform end-to-end emboied visual tracking. To ensure standardized evaluation, the maximum duration of each episode is set to 500 steps. To provide a comprehensive analysis, the experimental results are categorized and reported based on two primary target types: vehicles (including two-wheelers) and pedestrians. Detailed environmental parameters, including weather conditions and initialization offsets, are provided in the supplementary material.

EVT requires the UAV to maintain both semantic locking and physical following of the target in 3D space. A tracking state is considered valid only when the target’s bounding box remains within the camera’s Field of View (FOV). To assess the model’s fine-grained understanding of spatial prepositions in natural language instructions, we define three distance-based tracking criteria:close, suitable, and long. Specifically, for vehicles, the close, suitable, and long distances are defined as $\le$25m, $\le$35m, and $\le$40m, respectively. For pedestrians, owing to their smaller scale, these corresponding thresholds are inherently tighter, set at $\le$10m, $\le$15m, and $\le$20m.

The outcome of a tracking mission is determined by the following criteria: a mission is terminated and marked as a failure if an abnormal tracking state (i.e., violating FOV or distance constraints) persists for 15 consecutive frames. A task achieves success simply by surviving a 200-step episode without ever triggering this continuous failure condition.

UAV-Track VLA is optimized end-to-end using a joint loss function. As detailed in the methodology section, this joint objective aligns visual features with language instructions while supervising action execution. Specific weight configurations for each loss component are detailed in the supplementary material. The model is trained on two NVIDIA H100 GPUs with a global batch size of 64 for 45,000 iterations. We utilize absolute position changes as the primary supervision signal, enabling the flow-matching action expert to effectively fit the action flow field in a continuous state space.
\begin{table*}[t]
\centering
\caption{Quantitative results on Seen Maps}
\label{tab:seen_maps}
\resizebox{\textwidth}{!}{
\begin{tabular}{l c c c c c c c}
\toprule
\multirow{2}{*}{\textbf{Model}} & \multirow{2}{*}{\textbf{Target}} & \multicolumn{2}{c}{\textbf{Close}} & \multicolumn{2}{c}{\textbf{Suitable}} & \multicolumn{2}{c}{\textbf{Far}} \\
\cmidrule(lr){3-4} \cmidrule(lr){5-6} \cmidrule(lr){7-8}
 & & \textbf{ATF} & \textbf{SR (\%)} & \textbf{ATF} & \textbf{SR (\%)} & \textbf{ATF} & \textbf{SR (\%)} \\
\midrule
ACT\cite{zhao2023learningfinegrainedbimanualmanipulation} & Veh. / Ped. & 36.00 / 18.50 & 0.00 / 0.00 & 30.17 / 20.70 & 0.00 / 0.00 & 33.24 / 21.52 & 0.00 / 0.00 \\
WALL-OSS\cite{zhai2025ignitingvlmsembodiedspace} & Veh. / Ped. & 25.67 / 44.51 & 0.00 / 0.00 & 47.80 / 55.88 & 0.00 / 2.44 & 40.29 / 41.62 & 0.00 / 0.00 \\
$\pi_{0}$\cite{black2026pi0visionlanguageactionflowmodel} & Veh. / Ped. & 67.14 / 86.27 & 1.43 / 0.00 & 94.53 / 142.78 & 7.35 / 18.75 & 89.14 / 133.53 & 3.39 / 17.65 \\
$\pi_{0.5}$\cite{intelligence2025pi05visionlanguageactionmodelopenworld} & Veh. / Ped. & 81.07 / 110.44 & 5.56 / 7.41 & 138.55 / 191.97 & 22.54 / 28.57 & 174.81 / 214.28 & 33.90 / 44.44 \\
\midrule
\textbf{UAV-Track VLA (Ours)} & Veh. / Ped. & \textbf{98.47} / \textbf{203.93} & \textbf{12.73} / \textbf{31.82} & \textbf{164.79} / \textbf{263.61} & \textbf{29.61} / \textbf{60.20} & \textbf{194.91} / \textbf{269.65} & \textbf{37.88} / \textbf{61.76} \\
\bottomrule
\end{tabular}
}
\end{table*}

\begin{table*}[t]
\centering
\caption{Zero-shot performance on Unseen Maps}
\label{tab:unseen_maps}
\resizebox{\textwidth}{!}{
\begin{tabular}{l c c c c c c c}
\toprule
\multirow{2}{*}{\textbf{Model}} & \multirow{2}{*}{\textbf{Target}} & \multicolumn{2}{c}{\textbf{Close}} & \multicolumn{2}{c}{\textbf{Suitable}} & \multicolumn{2}{c}{\textbf{Far}} \\
\cmidrule(lr){3-4} \cmidrule(lr){5-6} \cmidrule(lr){7-8}
 & & \textbf{ATF} & \textbf{SR (\%)} & \textbf{ATF} & \textbf{SR (\%)} & \textbf{ATF} & \textbf{SR (\%)} \\
\midrule
ACT\cite{zhao2023learningfinegrainedbimanualmanipulation} & Veh. / Ped. & 31.21 / 18.35 & 0.00 / 0.00 & 31.61 / 18.65 & 0.00 / 0.00 & 39.14 / 19.71 & 0.00 / 0.00 \\
WALL-OSS\cite{zhai2025ignitingvlmsembodiedspace} & Veh. / Ped. & 29.63 / 39.30 & 0.00 / 0.00 & 46.59 / 52.56 & 0.00 / 0.00 & 34.78 / 41.82 & 0.00 / 0.00 \\
$\pi_{0}$\cite{black2026pi0visionlanguageactionflowmodel} & Veh. / Ped. & 76.05 / 84.83 & 2.63 / 4.35 & 95.22 / 148.20 & 10.00 / 20.00 & 76.91 / 146.42 & 2.22 / 21.05 \\
$\pi_{0.5}$\cite{intelligence2025pi05visionlanguageactionmodelopenworld} & Veh. / Ped. & \textbf{83.49} / 130.86 & 4.88 / 23.81 & \textbf{201.72} / 135.38 & \textbf{33.33} / 23.81 & 128.20 / 139.18 & 17.07 / 5.88 \\
\midrule
\textbf{UAV-Track VLA (Ours)} & Veh. / Ped. & 75.56 / \textbf{148.94} & \textbf{6.67} / \textbf{29.41} & 150.00 / \textbf{210.24} & 29.79 / \textbf{36.84} & \textbf{159.84} / \textbf{226.90} & \textbf{27.91} / \textbf{55.00} \\
\bottomrule
\end{tabular}
}
\end{table*}

\begin{table*}[t]
\centering
\caption{Ablation Study: Impact of the Spatial Auxiliary Head}
\label{tab:ablation}
\resizebox{\textwidth}{!}{
\begin{tabular}{l l c c c c c c c}
\toprule
\multirow{2}{*}{\textbf{Map Type}} & \multirow{2}{*}{\textbf{Setting}} & \multirow{2}{*}{\textbf{Target}} & \multicolumn{2}{c}{\textbf{Close}} & \multicolumn{2}{c}{\textbf{Suitable}} & \multicolumn{2}{c}{\textbf{Far}} \\
\cmidrule(lr){4-5} \cmidrule(lr){6-7} \cmidrule(lr){8-9}
 & & & \textbf{ATF} & \textbf{SR (\%)} & \textbf{ATF} & \textbf{SR (\%)} & \textbf{ATF} & \textbf{SR (\%)} \\
\midrule
\multirow{2}{*}{Seen} & w/o Aux Head & Veh. / Ped. & 80.45 / 125.43 & 10.45 / 10.00 & \textbf{176.13} / 166.67 & \textbf{29.87} / 23.33 & 161.76 / 183.96 & 30.88 / 39.29 \\
 & \textbf{UAV-Track VLA} & Veh. / Ped. & \textbf{98.47} / \textbf{203.93} & \textbf{12.73} / \textbf{31.82} & 164.79 / \textbf{263.61} & 29.61 / \textbf{60.20} & \textbf{194.91} / \textbf{269.65} & \textbf{37.88} / \textbf{61.76} \\
\midrule
\multirow{2}{*}{Unseen} & w/o Aux Head & Veh. / Ped. & 70.22 / 96.96 & 3.12 / 12.50 & \textbf{174.24} / 137.48 & 26.83 / 8.70 & 120.98 / 159.22 & 16.67 / 22.22 \\
 & \textbf{UAV-Track VLA} & Veh. / Ped. & \textbf{75.56} / \textbf{148.94} & \textbf{6.67} / \textbf{29.41} & 150.00 / \textbf{210.24} & \textbf{29.79} / \textbf{36.84} & \textbf{159.84} / \textbf{226.90} & \textbf{27.91} / \textbf{55.00} \\
\bottomrule
\end{tabular}
}
\end{table*}

\subsection{Quantitative Results and Analysis}
We compare UAV-Track VLA against several baselines, including $\pi_{0}$, the original $\pi_{0.5}$, and traditional control-based methods. Average results across multiple towns are summarized in Tables~\ref{tab:seen_maps} and~\ref{tab:unseen_maps}.

In seen maps (Town02, Town05, Town06, Town07, Town10HD), UAV-Track VLA demonstrates superior adaptability across diverse target types and motion patterns. As shown in Table~\ref{tab:seen_maps}, our model achieves high success rates (SR) and average tracking steps. For instance, in the "Far" pedestrian tracking task, UAV-Track VLA achieves an average of 269.65 steps with a 61.76\% SR, significantly outperforming all baseline models. For the vehicle tracking task under the same distance constraint, the average steps reach 194.91 with a 37.88\% SR.

To evaluate robustness, we deploy the model in entirely unseen CARLA towns (Town01, Town03, Town04). Results in Table~\ref{tab:unseen_maps} indicate that UAV-Track VLA maintains stable performance even without environment priors. For pedestrian tracking at "Suitable" and "Far" distances, the model retains an average of 210.24 and 226.90 steps, respectively, with SRs only slightly lower than those in seen maps. These results validate the model's effective performance and strong generalization ability in unseen scenes.

A key observation is that while UAV-Track VLA performs comparably to the original $\pi_{0.5}$ on vehicle tracking, it shows a substantial lead in pedestrian tracking. In unseen maps ("Far" distance), $\pi_{0.5}$'s SR for pedestrians drops to 5.88\%, whereas UAV-Track VLA maintains 55.00\%. This clearly demonstrates the significant advantage of our improved model when handling pedestrian targets compared to vehicle targets.

\begin{table}[h]
\centering
\caption{Inference Efficiency Comparison}
\label{tab:latency}
\begin{tabular}{l c}
\toprule
\textbf{Model} & \textbf{Average Latency (s)} \\
\midrule
WALL-OSS\cite{zhai2025ignitingvlmsembodiedspace} & 0.4524 \\
$\pi_{0.5}$\cite{intelligence2025pi05visionlanguageactionmodelopenworld} & 0.0857 \\
$\pi_{0}$\cite{black2026pi0visionlanguageactionflowmodel} & 0.0691 \\
\midrule
\textbf{UAV-Track VLA (Ours)} & \textbf{0.0571} \\
\bottomrule
\end{tabular}
\end{table}

\subsection{Ablation Analysis}
\subsubsection{Effectiveness of the Auxiliary Head.} 
We investigate the contribution of the spatial auxiliary head by comparing the full UAV-Track VLA against a variant without this module (denoted as \textit{w/o Aux Head} in Table~\ref{tab:ablation}). The absence of the auxiliary head leads to a severe performance decay in pedestrian tracking tasks. In seen maps ("Far"), the average steps drop from 269.65 to 183.96, and the SR falls from 61.76\% to 37.88\%. This performance gap is even more pronounced in unseen maps, dropping from 226.90 steps to 159.22 steps. These results demonstrate that the auxiliary head is essential for achieving precise vision-language alignment and spatial distance estimation, which is critical for maintaining accurate distances during pedestrian tracking.

\subsubsection{Inference Efficiency.} 
To evaluate the real-time control capabilities, we conduct a timing analysis of the single-step continuous inference latency across different models. To ensure a fair and meaningful evaluation, our comparison focuses strictly on models with similar parameter scales. Consequently, the ACT model is excluded from this analysis due to its significantly smaller parameter size and distinct architecture. Based on a statistical analysis of approximately 1,200 valid data points per model, the average inference latency is summarized in Table~\ref{tab:latency}. The results reveal that while the original $\pi_{0.5}$ requires 0.0857s per step, our UAV-Track VLA reduces the average latency to 0.0571s, achieving a substantial 33.4\% improvement in computational efficiency. This efficiency gain is primarily attributed to the temporal compression net, which effectively eliminates the redundancy of multi-frame inputs while preserving essential temporal context for high-precision tracking.

\section{Conclusion}
In this paper, we presented UAV-Track VLA, a novel end-to-end embodied visual tracking framework designed to tackle the highly dynamic and complex nature of urban UAV tracking. To bridge the gap between high-level language semantics and high-frequency low-level flight control, we introduced a parallel dual-branch decoding architecture that effectively decouples spatial grounding from continuous action generation. By incorporating a temporal compression net and a spatial-aware auxiliary grounding head, our model efficiently extracts dynamic evolution patterns and injects precise geometric priors into the flow matching action expert. Furthermore, we proposed UAV-Track, a large-scale multimodal benchmark dedicated to natural language-guided emboied UAV tracking, comprising over 890K frames of highly diverse simulated data. Extensive experiments in the CARLA simulator demonstrate that our approach significantly outperforms existing baselines in tracking robustness, cross-modal alignment, and zero-shot generalization. Notably, our framework excels in complex pedestrian tracking scenarios while achieving a 33.4\% reduction in single-step inference latency. In future work, we plan to deploy our framework on physical UAV platforms to further investigate sim-to-real transfer and real-world embodied tracking capabilities.

\bibliographystyle{unsrt}  
\bibliography{references}  
\clearpage
\appendix
\section*{Appendix}
\section{UAV-Track Benchmark Details}

\subsection{Data Collection and Environment Configurations}

\subsubsection{Sensor and Asynchronous Sampling Settings}
To acquire high-fidelity visual observations and control signals, a forward-facing RGB camera is mounted $0.5$ m directly below the UAV's center of mass $(x=0, y=0, z=-0.5)$. Its relative pose is fixed at $\text{yaw}=0^{\circ}$, $\text{roll}=0^{\circ}$, and $\text{pitch}=-15^{\circ}$. The camera resolution is set to $800 \times 600$ pixels with a $135^{\circ}$ Field of View (FOV). All simulations are conducted in CARLA 0.9.14 powered by Unreal Engine 4.27.

To balance computational efficiency and control smoothness, we adopt an asynchronous sampling strategy. Visual observations are captured at 5 Hz to extract macroscopic motion trends while avoiding temporal feature redundancy. Concurrently, the UAV's proprioceptive states and action chunking sequences are recorded at 25 Hz to support high-frequency, continuous displacement flow matching.

\subsubsection{Global Environment and Dynamic Distractors}
To mitigate the sim-to-real domain shift, we introduce extensive domain randomization at the onset of each episode. Weather and illumination are randomized via \texttt{carla.WeatherParameters}, uniformly sampling cloudiness $[0, 35]$, precipitation $[0, 40]$, precipitation deposits $[0, 30]$, wind intensity $[0, 10]$, fog density $[0, 30]$, fog distance $[100, 200]$ m, wetness $[0, 10]$, sun azimuth $[0^{\circ}, 360^{\circ}]$, and sun altitude $[-5^{\circ}, 90^{\circ}]$. The lower bound of $-5^{\circ}$ for sun altitude effectively simulates extremely low-light nighttime tracking scenarios. Furthermore, 30 dynamic vehicles (including two-wheelers) and 20 roaming pedestrians are spawned per episode. These entities serve as candidate targets as well as highly dynamic background distractors and obstacles, compelling the model to learn robust, interference-invariant representations.

\subsubsection{Expert Demonstrations and Control Granularity}
The dataset encompasses approximately 200K frames of manual expert demonstrations. To accommodate the distinct kinematic characteristics of heterogeneous targets, we enforce differential control granularity. The step sizes for vertical (Z-axis) displacement and yaw angle are fixed at $0.35$ m and $3.5^{\circ}$, respectively. For horizontal tracking (X and Y axes), the control resolution is set to $1.35$ m for high-speed vehicles, whereas it is refined to $0.1$ m for slow-moving pedestrians to facilitate smooth, close-range micro-adjustments.

\subsubsection{Automated Exploration and APF-based Obstacle Avoidance}
To alleviate the covariate shift inherent in end-to-end imitation learning, we design an automated data collection paradigm integrating random perturbations with an Artificial Potential Field (APF). Initially, the UAV is randomly initialized within a predefined spatial boundary behind the target (detailed in Table~\ref{tab:init_params}). During automated cruising, continuous random noise is injected into the 4-DoF actions, coupled with a regression term (coefficient 0.6) that compels the UAV to continuously attempt to restore its initial relative anchor pose.

When obstacles are detected within safety margins ($0.35$ m for pedestrians, $0.5$ m for vehicles, and $0.15$ m for others), the APF mechanism is activated. It suspends random perturbations and applies lateral and longitudinal repulsive forces (maximum magnitude 1.5) along with an anti-ground collision force (ensuring a minimum altitude of $0.1$ m) to guide the UAV in collision evasion.

\begin{table}[htbp]
\centering
\caption{Spatial Configurations for Automated Data Collection.}
\label{tab:init_params}
\resizebox{\textwidth}{!}{
\begin{tabular}{llcccc}
\toprule
\textbf{Target} & \textbf{Distance} & \textbf{X-axis Range (m)} & \textbf{Y-axis Range (m)} & \textbf{Z-offset Range (m)} & \textbf{Yaw Perturbation Range (Rear/Right-Rear/Left-Rear)} \\
\midrule
\multirow{3}{*}{Vehicle} & Near & $[-2.5, -1.75]$ & $[-3.0, 3.0]$ & $[0.2, 1.5]$ & $[-10^{\circ}, 10^{\circ}]$ / $[-60^{\circ}, -10^{\circ}]$ / $[10^{\circ}, 60^{\circ}]$ \\
 & Close & $[-4.0, -1.0]$ & $[-2.0, 2.0]$ & $[0.3, 2.5]$ & $[-10^{\circ}, 10^{\circ}]$ / $[-60^{\circ}, -10^{\circ}]$ / $[10^{\circ}, 60^{\circ}]$ \\
 & Far & $[-8.0, -2.0]$ & $[-1.5, 1.5]$ & $[0.6, 5.0]$ & $[-10^{\circ}, 10^{\circ}]$ / $[-60^{\circ}, -10^{\circ}]$ / $[10^{\circ}, 60^{\circ}]$ \\
\midrule
\multirow{3}{*}{Pedestrian} & Near & $[-1.5, -0.5]$ & $[-0.5, 0.5]$ & $[0.1, 0.7]$ & $[-8^{\circ}, 8^{\circ}]$ / $[-50^{\circ}, -8^{\circ}]$ / $[8^{\circ}, 50^{\circ}]$ \\
 & Close & $[-2.0, -0.8]$ & $[-1.0, 1.0]$ & $[0.2, 1.5]$ & $[-8^{\circ}, 8^{\circ}]$ / $[-50^{\circ}, -8^{\circ}]$ / $[8^{\circ}, 50^{\circ}]$ \\
 & Far & $[-4.0, -1.6]$ & $[-0.8, 0.8]$ & $[0.4, 3.0]$ & $[-8^{\circ}, 8^{\circ}]$ / $[-50^{\circ}, -8^{\circ}]$ / $[8^{\circ}, 50^{\circ}]$ \\
\bottomrule
\end{tabular}
}
\end{table}
\begin{figure}[htbp]
    \centering
    \includegraphics[width=0.9\textwidth]{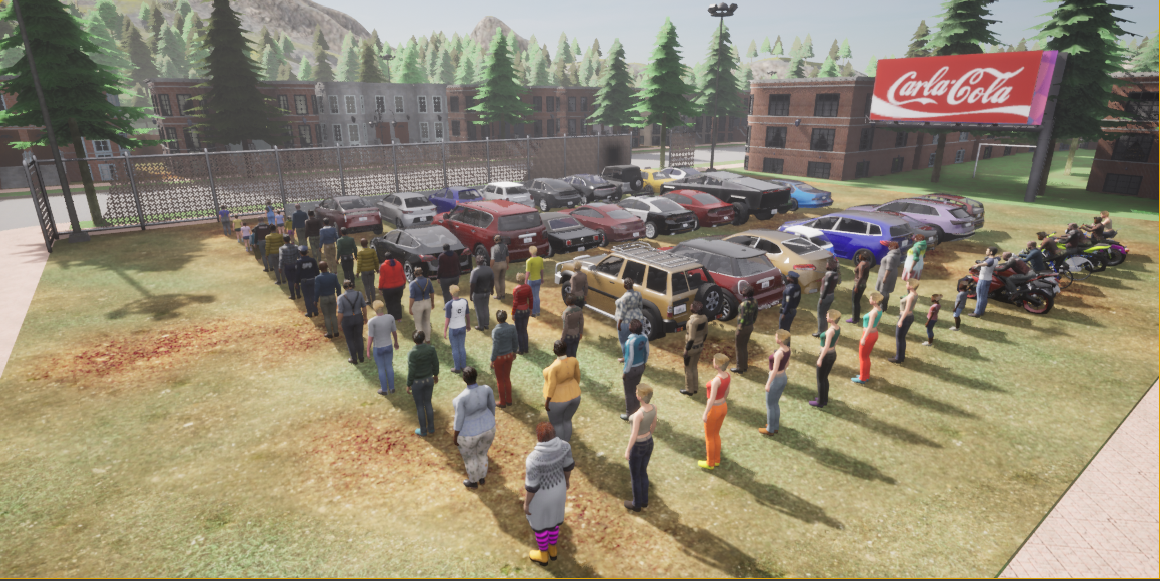}
    \caption{\textbf{Exemplar Gallery of Heterogeneous Tracking Targets.} This collage provides a visual overview of the diverse candidate assets utilized as tracking targets within the UAV-Track benchmark. The targets encompass standard vehicles, two-wheelers (bicycles, motorcycles), and pedestrians with varying demographics (adults, children). These assets form the basis for the fine-grained semantic attribute mapping (e.g., color, age) used in the instructions described in Section A.2.}
    \label{fig:target_assets}
\end{figure}

\subsection{Data Processing and Prompt Generation}

\subsubsection{Egocentric Local Coordinate Alignment}
To preclude the model from learning spurious correlations associated with global reference frames, we rigorously transform all recorded positional data into a UAV-centric relative coordinate system. Specifically, given the current 6-DoF camera pose $(x, y, z, \text{roll}, \text{yaw}, \text{pitch})$, we translate the global origin to the camera center and construct a composite rotation matrix $R_{\text{world\_to\_camera}}$ via Euler angles. All 3D coordinates are subsequently batch-projected into this relative space, ensuring the policy strictly relies on egocentric visual features for spatial reasoning.

\subsubsection{Visual Pre-processing}
To maintain consistency with the foundational vision-language models and prevent the distortion of spatial geometric priors, we adopt a rigid visual pre-processing strategy identical to $\pi_{0.5}$. The original high-resolution $800 \times 600$ images are downscaled to fit a $224 \times 224$ input resolution. Instead of direct resizing, which would distort the aspect ratio, the image is proportionally scaled based on the larger dimension (width reduction from 800 to 224). Subsequently, zero-padding is applied to the height dimension. This operation ensures that the aspect ratio remains perfectly intact, which is critical for the spatial-aware auxiliary head to accurately estimate the 3D relative distance and yaw angle.

\subsubsection{Fine-Grained Target Attribute Mapping}
We construct a comprehensive vocabulary of 176 natural language tracking instructions, structured as \texttt{[Verb] + [Target Attribute] + [Distance Constraint]}. To map simulated parameters to human-readable language, we employ a fine-grained semantic mapping strategy:

\textbf{Vehicle Colors:} The native RGB hex codes from CARLA are extracted. Based on luminance, base tones (black, white, gray) are filtered first. For colored vehicles, the Euclidean distance in the RGB space is calculated against 18 predefined natural language colors to assign the closest semantic label.

\textbf{Pedestrian Attributes:} Pedestrians are semantically tagged utilizing a combination of age brackets (Adult/Teenager/Child) and gender (Male/Female).

\subsubsection{Controlled Zero-Shot Prompt Generation}
The 176 instructions are split into 136 seen training prompts and 40 unseen testing prompts. To prevent drastic shifts in task physics caused by abrupt instruction alterations, we propose a \textit{controlled single-variable substitution} strategy. During the generation of an unseen prompt, exactly one semantic block from the base instruction is randomly substituted with a synonym.

\subsection{Tracking Instructions}
To ensure transparency and support future reproducibility, we detail the composition of the 176 natural language instructions used in the UAV-Track benchmark.

\subsubsection{Training Prompts (Seen Dataset)}
The 136 seen instructions in the training set are generated through a combinatorial template mapping. The base template is: \texttt{Track the [Target Attribute] at a [Distance Constraint] distance}. The complete list of 136 training prompts is available in our open-source repository.

\subsubsection{Zero-Shot Testing Prompts (Unseen Dataset)}
The 40 unseen prompts used for zero-shot testing are detailed in Table~\ref{tab:zero_shot_prompts}. These prompts force the model to interpret semantic equivalents and altered operational intents without prior training exposure.

\subsection{Dataset Organization and Formats}
To facilitate broad adoption by the research community and ensure seamless integration with various embodied AI training pipelines, the UAV-Track dataset is publicly released in two standardized formats: HDF5 and LeRobot V2.1 (Parquet).

\subsubsection{HDF5 Format}
The HDF5 version organizes data using a flat directory structure, where each tracking trajectory is stored as an independent file (e.g., \texttt{episode\_0.hdf5}). This format is highly accessible for custom dataloaders and step-by-step debugging. Within each file, the multimodal data is hierarchically structured into groups and datasets:

\textbf{\texttt{Observations/Images/}}: Contains the current front-view RGB frame (\texttt{cam\_high}) and the three historical frames (\texttt{frame(i-1)} to \texttt{frame(i-3)}). Each frame is stored as a \texttt{uint8} array with a shape of $600 \times 800 \times 3$.

\textbf{\texttt{State}}: A \texttt{float32} array recording the UAV's proprioceptive observations at each step. The 4-dimensional vector explicitly represents the 3D linear velocities and the yaw angular velocity.

\textbf{\texttt{Prompts}}: An array of strings storing the specific natural language instruction assigned to the episode.

\textbf{\texttt{Action}}: A \texttt{float32} array storing the continuous future action sequences (i.e., the future positions and yaw angles for the subsequent 25 steps) corresponding to each observation.

\subsubsection{LeRobot V2.1 Format}
For highly optimized, large-scale data loading and native compatibility with the Hugging Face ecosystem, we also provide the dataset strictly formatted according to the LeRobot V2.1 standard. The directory is cleanly partitioned into data chunks and global metadata:

\begin{verbatim}
all_data_V1.0/
|-- data/
|   |-- chunk-000/
|   |   |-- episode_000000.parquet
|   |   |-- episode_000001.parquet
|   |   `-- ...
|   `-- chunk-001/
|       `-- ...
|-- meta/
|   |-- episodes.jsonl
|   |-- episodes_stats.jsonl
|   |-- info.json
|   `-- tasks.jsonl
\end{verbatim}

The \texttt{meta/} directory aggregates dataset-level statistics, task-index mappings, and episode lengths, enabling rapid filtering without loading heavy payloads. Each \texttt{.parquet} file within the \texttt{data/} chunks follows a strict columnar schema designed for efficient parallel batching:

 \textbf{\texttt{Observation.Images}}: Structs containing binary image bytes and relative storage paths for both current and historical visual observations.

\textbf{\texttt{Observation.State}}: A fixed-size list (\texttt{float32[4]}) representing the kinematic velocity inputs.

\textbf{\texttt{Action}}: A 2D array of \texttt{float32} mapped to the $[25 \times 4]$ future trajectory displacements predicted by the flow matching action expert.

\textbf{\texttt{Position}}: The absolute spatial 3D coordinates and yaw of the UAV, provided for potential downstream metric calculations.

\textbf{Metadata}: Essential index columns including \texttt{timestamp}, \texttt{frame\_index}, \texttt{episode\_index}, and \texttt{task\_index} are natively embedded to support temporal sequence reconstruction and task-conditioned training.

\begin{table}[htbp]
\centering
\caption{Detailed Mapping of Unseen Zero-Shot Prompts against Seen Training Prompts}
\label{tab:zero_shot_prompts}
\resizebox{\textwidth}{!}{
\begin{tabular}{cll}
\toprule
\textbf{Substitution Type} & \textbf{Original Base Prompt (Seen)} & \textbf{Unseen Test Prompt (Zero-Shot)} \\
\midrule
\multirow{14}{*}{\textbf{Verb Changed}} 
& Track the child male pedestrian at a long distance. & Focus on the child male pedestrian at a long distance. \\
& Track the red bicycle at a long distance. & Keep an eye on the red bicycle at a long distance. \\
& Track the vehicle at a long distance. & Focus on the vehicle at a long distance. \\
& Track the child female pedestrian at a long distance. & Focus on the child female pedestrian at a long distance. \\
& Track the black motorcycle at a suitable distance. & Pursue the black motorcycle at a suitable distance. \\
& Track the orange motorcycle at a long distance. & Pursue the orange motorcycle at a long distance. \\
& Track the orange motorcycle at a suitable distance. & Pursue the orange motorcycle at a suitable distance. \\
& Track the red bicycle at a suitable distance. & Focus on the red bicycle at a suitable distance. \\
& Track the child female pedestrian at a suitable distance. & Focus on the child female pedestrian at a suitable distance. \\
& Track the child male pedestrian at a suitable distance. & Keep an eye on the child male pedestrian at a suitable distance. \\
& Track the black motorcycle at a long distance. & Pursue the black motorcycle at a long distance. \\
& Track the black vehicle at a suitable distance. & Pursue the black vehicle at a suitable distance. \\
& Track the pedestrian at a long distance. & Keep an eye on the pedestrian at a long distance. \\
& Track the adult female pedestrian at a long distance. & Focus on the adult female pedestrian at a long distance. \\
\midrule
\multirow{17}{*}{\textbf{Object Changed}}
& Track the black vehicle at a long distance. & Track the black auto at a long distance. \\
& Track the adult male pedestrian at a long distance. & Track the male adult at a long distance. \\
& Track the green motorcycle at a long distance. & Track the green motorbike at a long distance. \\
& Track the child female pedestrian at a long distance. & Track the little girl at a long distance. \\
& Track the black motorcycle at a long distance. & Track the black motorbike at a long distance. \\
& Track the dark blue vehicle at a suitable distance. & Track the dark blue auto at a suitable distance. \\
& Track the pedestrian at a suitable distance. & Track the human at a suitable distance. \\
& Track the dark red vehicle at a suitable distance. & Track the dark red automobile at a suitable distance. \\
& Track the blue bicycle at a long distance. & Track the blue pedal cycle at a long distance. \\
& Track the pedestrian at a suitable distance. & Track the walker at a suitable distance. \\
& Track the light gray vehicle at a suitable distance. & Track the light gray auto at a suitable distance. \\
& Track the adult male pedestrian at a suitable distance. & Track the man at a suitable distance. \\
& Track the child male pedestrian at a suitable distance. & Track the boy at a suitable distance. \\
& Track the dark red vehicle at a suitable distance. & Track the dark red automobile at a suitable distance. \\
& Track the red bicycle at a long distance. & Track the red cycle at a long distance. \\
& Track the adult female pedestrian at a suitable distance. & Track the woman at a suitable distance. \\
& Track the child male pedestrian at a long distance. & Track the boy at a long distance. \\
\midrule
\multirow{9}{*}{\textbf{Distance Changed}}
& Track the adult female pedestrian at a suitable distance. & Track the adult female pedestrian nearby. \\
& Track the adult male pedestrian at a close distance. & Track the adult male pedestrian at a close range. \\
& Track the orange motorcycle at a suitable distance. & Track the orange motorcycle nearby. \\
& Track the blue bicycle at a suitable distance. & Track the blue bicycle nearby. \\
& Track the black vehicle at a close distance. & Track the black vehicle at a close range. \\
& Track the red bicycle at a suitable distance. & Track the red bicycle nearby. \\
& Track the dark gray motorcycle at a close distance. & Track the dark gray motorcycle at a close range. \\
& Track the dark red vehicle at a long distance. & Track the dark red vehicle from afar. \\
& Track the white vehicle at a suitable distance. & Track the white vehicle nearby. \\
\bottomrule
\end{tabular}
}
\end{table}
\subsection{Evaluation Metrics and Success Criteria}
To rigorously assess the closed-loop tracking performance, we introduce two core metrics: Success Rate (SR) and Average Tracked Frames (ATF).

\subsubsection{Success Criteria}
A mission is considered successful only if the system avoids a ``fatal failure'' throughout a 500-step evaluation episode. A fatal failure is triggered if the target continuously violates the camera's Field of View (FOV) or the defined distance constraints for a maximum tolerance threshold of 15 consecutive frames. It is important to note that success is defined by the absence of such fatal failures within the 500-step window, allowing for brief, recoverable tracking deviations as long as they do not exceed the tolerance threshold.
\subsubsection{Performance Metrics}
~\\ 
\indent \textbf{Success Rate (SR):} Defined as the ratio of successful episodes to the total number of test episodes $N$.

\textbf{Average Tracked Frames (ATF):} Measures the average duration of maintaining effective tracking (i.e., within FOV and range) before a fatal failure occurs. This metric provides a fine-grained evaluation of control stability under complex dynamics.

\section{UAV-Track VLA Architecture and Training Details}

\subsection{Architectural Specifics}
To facilitate reproducibility, we detail the specific architectural parameters of our proposed modules:

\textbf{Temporal Compression Net:} A bias-free linear projection layer is employed to drastically reduce the sequence length of historical visual features. It compresses the original 256 visual tokens of each of the 3 historical frames down to a compact representation of 64 tokens per frame.

\textbf{Visual Position Embedding:} After concatenating the current frame tokens with the compressed historical tokens, the fused sequence contains a total of 448 visual tokens (i.e., $3 \times 64 + 256$). A learnable positional embedding tensor of size $1 \times 448 \times D$ (where $D$ is the hidden size of the LLM) is added to this sequence. Crucially, we initialize this positional embedding strictly with constant zeros (\texttt{0.0}). This zero-initialization strategy ensures that the pre-trained visual feature space remains perfectly unperturbed at the onset of training, preventing early divergence while allowing the model to smoothly adapt to temporal dynamics.

 \textbf{Spatial-Aware Auxiliary Grounding Head:} This module is instantiated as a cross-attention pooling layer followed by a Multi-Layer Perceptron (MLP). A learnable query vector, initialized from a standard normal distribution ($\sigma=0.02$), attends to the cross-modal hidden states. The resulting pooled representation is then processed by a two-layer MLP utilizing a GELU activation function to output the 4D spatial relative pose. All linear projection weights within this module employ Xavier uniform initialization, with biases initialized to zero.

\subsection{Training Hyperparameters and Optimization}
Unlike parameter-efficient tuning methods (e.g., LoRA), we employ \textbf{full fine-tuning} across all model parameters to maximize the VLA model's adaptation to the aerial continuous control space. The end-to-end joint loss function is configured with weight coefficients $\lambda_1 = 2$ (position loss) and $\lambda_2 = 0.1$ (flow matching action loss). 

The network is optimized using the AdamW optimizer with a gradient clipping norm of 0.8 to stabilize the training of the flow matching expert. We utilize a Cosine Decay learning rate schedule: the learning rate warms up to a peak of $1.2 \times 10^{-4}$ over the first 2,000 steps, and subsequently decays to $1.0 \times 10^{-6}$ over the remaining 35,000 steps. Furthermore, an Exponential Moving Average (EMA) with a decay rate of 0.999 is applied to the model weights to ensure smooth convergence and enhance inference robustness.

\subsection{Baselines Adaptation Details}
To guarantee a fair and rigorous comparative analysis, all baseline models evaluated in our benchmark are strictly aligned to output the identical 4-DoF continuous action space (3D displacement and yaw offset) with a temporal chunking size of 25 steps. 
Specifically, for the foundational VLA baselines, $\pi_0$ and $\pi_{0.5}$, we utilize their official open-source repositories, modifying the action decoders and applying full fine-tuning on our UAV-Track dataset. For the traditional embodied control baselines, ACT and WALL-OSS, we leverage the highly-optimized implementations provided by the Hugging Face LeRobot repository. Both models undergo full fine-tuning and inference under the same 25-step prediction horizon constraints.

\section{Additional Experiment}
\begin{figure}[htbp]
    \centering
    \includegraphics[width=\textwidth]{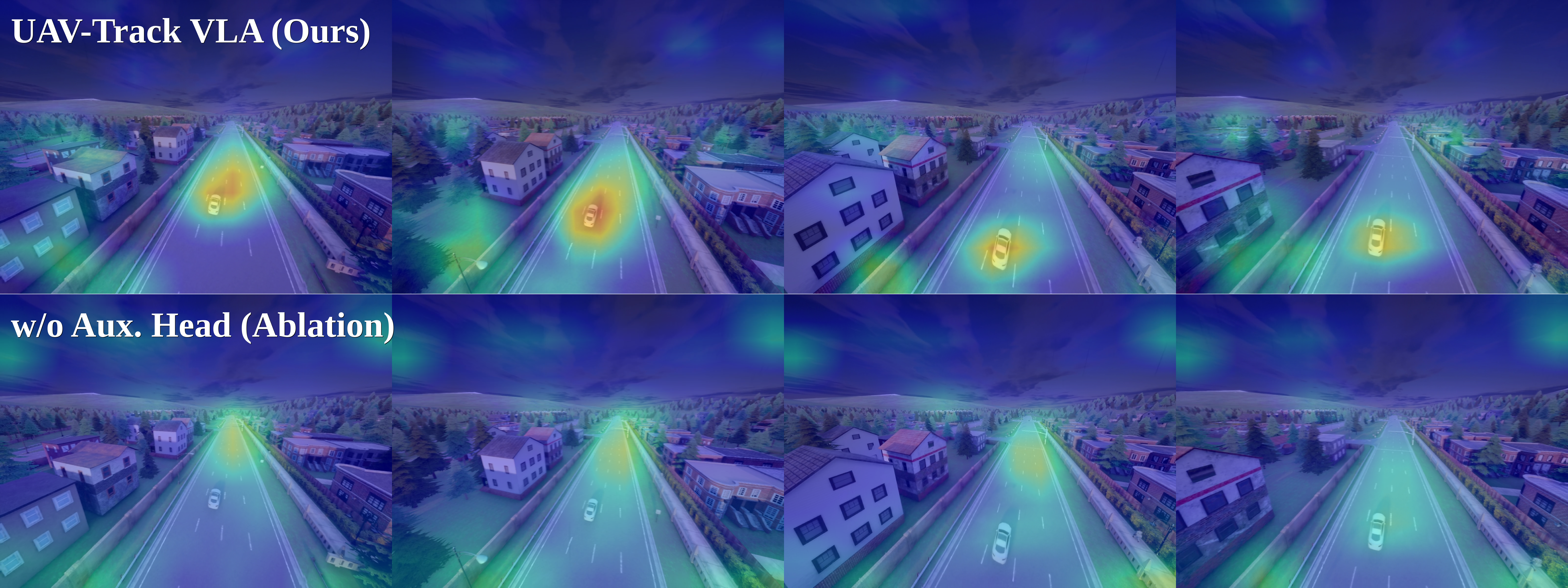}
    \includegraphics[width=\textwidth]{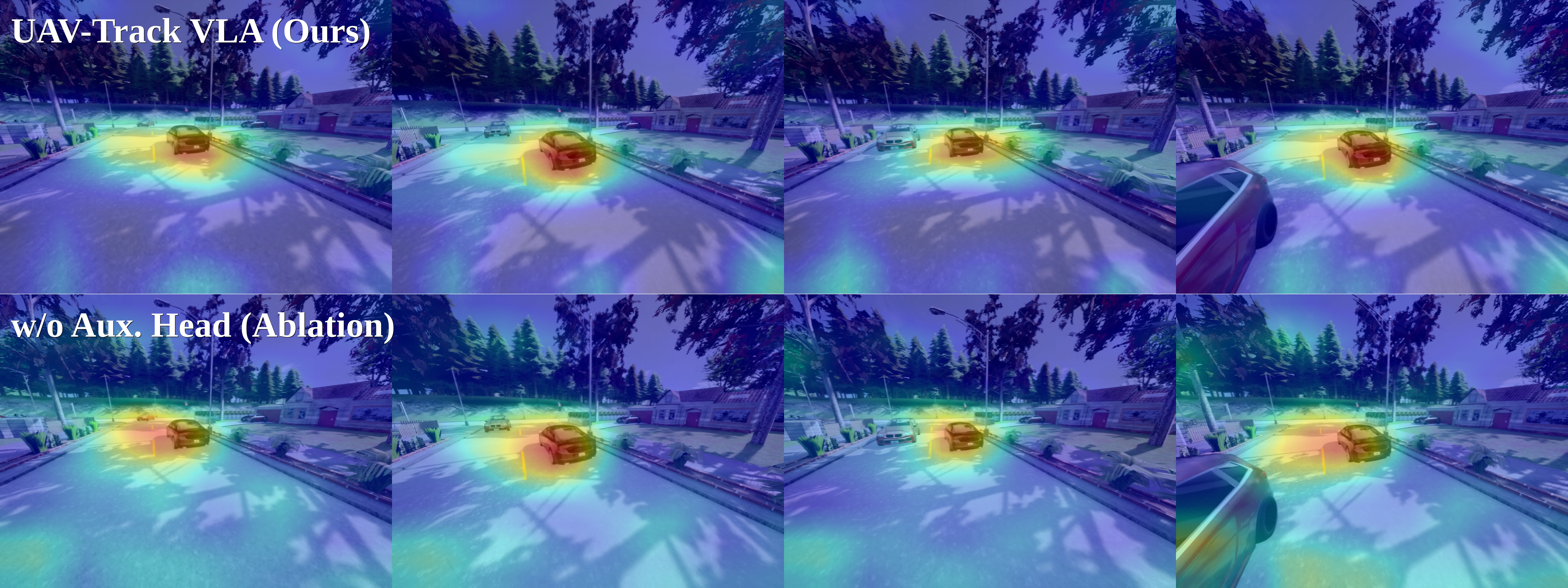}
    \caption{\textbf{Visualization of Cross-Modal Attention Maps.} We visualize the average attention weights from language tokens to the current frame's visual tokens across the last four layers of the LLM backbone. Top rows: The full UAV-Track VLA model exhibits highly concentrated and accurate attention on the specific target. Bottom rows: The ablated model without the Spatial-Aware Auxiliary Grounding Head shows diffuse and scattered attention, struggling to ground the semantic instruction in the visual space.}
    \label{fig:attention_vis}
\end{figure}

\subsection{Qualitative Analysis of Cross-Modal Alignment}

To further investigate the underlying mechanism by which the Spatial-Aware Auxiliary Grounding Head enhances tracking performance, we conduct a qualitative analysis of the cross-modal attention maps within the network. Specifically, we extract the attention weights mapping the language instruction tokens to the visual tokens of the current observation frame. To obtain a stable and robust representation of the model's focus, we compute the average of these attention scores across the final four layers of the LLM backbone and project them as heatmaps onto the original RGB images.

As illustrated in Figure~\ref{fig:attention_vis}, we compare the attention distributions generated by our full UAV-Track VLA model (top rows) against the ablated variant without the auxiliary head (bottom rows) across different tracking scenarios. The visualizations reveal a stark contrast in semantic grounding capability.

In the full UAV-Track VLA model, the attention is highly localized and sharply concentrated on the correct target vehicle. The model successfully filters out complex background distractors, such as roads, vegetation, and irrelevant adjacent vehicles. Conversely, the ablated model exhibits highly diffuse and scattered attention weights. Without the explicit geometric priors provided by the auxiliary head, the LLM backbone struggles to precisely align the semantic language instruction with the corresponding physical instance in the continuous 3D visual space. 

This qualitative evidence strongly corroborates our quantitative ablation findings. It demonstrates that the auxiliary spatial supervision does not merely serve as an isolated geometric constraint; rather, its gradients propagate backward to fundamentally enhance the LLM backbone's intrinsic capacity for cross-modal grounding. By explicitly enforcing visual-semantic alignment, the auxiliary head ensures that the downstream flow-matching action expert receives accurately grounded features, which is indispensable for robust and highly dynamic aerial tracking.

\subsection{Prompt Sensitivity Analysis}

To investigate which semantic component of the natural language instruction exerts the most critical influence on the model's tracking stability, we conduct a sensitivity analysis utilizing the zero-shot testing prompts (as defined in Section B.3.2). The unseen prompts are categorized into three distinct intervention groups based on the single-variable substitution strategy: \textit{Object Changed}, \textit{Verb Changed}, and \textit{Distance Changed}.

\subsubsection{Normalization Methodology}
Given the inherent difficulty variations across different CARLA towns, directly averaging the raw metrics (Average Tracked Frames and Success Rate) may obscure the relative impact of each prompt component. Therefore, we adopt an intra-map normalization approach. For each specific map, the lowest performance among the three substitution types is defined as the baseline with a value of $1.0$. The performance metrics of the other substitution types are proportionally scaled relative to this worst-case baseline. Finally, these normalized scores are averaged across all evaluated towns to provide a global sensitivity metric.

\subsubsection{Results and Discussion}
The averaged normalized results are presented in Table~\ref{tab:prompt_sensitivity}. In this normalized scale, a lower value (closer to 1.0) indicates a more severe performance degradation, thereby implying that the model is highly sensitive to modifications in that specific prompt component.

\begin{table}[htbp]
\centering
\caption{Normalized Performance Impact of Single-Variable Prompt Substitution. Lower scores denote higher sensitivity (i.e., more severe performance drop when the component is altered).}
\label{tab:prompt_sensitivity}
\begin{tabular}{lcccc}
\toprule
\multirow{2}{*}{\textbf{Substituted Component}} & \multicolumn{2}{c}{\textbf{Vehicle Tracking}} & \multicolumn{2}{c}{\textbf{Pedestrian Tracking}} \\
\cmidrule(lr){2-3} \cmidrule(lr){4-5}
& \textbf{Norm. ATF} & \textbf{Norm. SR} & \textbf{Norm. ATF} & \textbf{Norm. SR} \\
\midrule
Object & 1.46 & 1.60 & \textbf{1.04} & \textbf{0.79} \\
Verb & \textbf{1.30} & \textbf{1.38} & 2.53 & 1.46 \\
Distance & 1.96 & 2.18 & 2.39 & 4.76 \\
\bottomrule
\end{tabular}
\end{table}

Based on the quantitative outcomes, we summarize the model's sensitivity to different prompt components as follows:

\textbf{High Sensitivity to Object Descriptions:} Overall, the model is most vulnerable to modifications in target object descriptions (e.g., substituting ``vehicle'' with ``car'', or ``female child pedestrian'' with ``girl''). Altering these nominal semantics results in the most severe performance drops across the evaluated maps, indicating that the model's visual locking is highly dependent on specific target phrasing.

\textbf{Moderate Impact of Action Verbs:} Changing the instructional verbs (e.g., swapping ``Track'' with ``Focus on'') leads to a moderate decrease in tracking stability. This demonstrates that the model maintains a generalized understanding of the tracking intent, though it still exhibits slight sensitivity to verb variations.

\textbf{Strong Robustness to Distance Constraints:} The model is consistently least sensitive to changes in distance adverbs (e.g., changing ``at a long distance'' to ``from afar''). The notably high normalized retention scores imply that variations in distance phrasing have a minimal impact on the fundamental tracking capabilities of the UAV-Track VLA model.

\subsection{Detailed Per-Town Performance}

In the main text, we reported the overall averaged tracking performance across seen and unseen environments. To provide a comprehensive and transparent evaluation, this section presents the disaggregated, per-map quantitative results. Tables \ref{tab:perf_town02} through \ref{tab:perf_town10} detail the model performances on the five seen maps during training, while Tables \ref{tab:perf_town01} through \ref{tab:perf_town04} report the zero-shot generalization performances on the three completely unseen maps. Best performances are highlighted in \textbf{bold}.


\begin{table}[htbp]
\centering
\caption{Detailed Tracking Performance on \textbf{Town02} (Seen Map).}
\label{tab:perf_town02}
\resizebox{\textwidth}{!}{
\begin{tabular}{l cc cc cc cc cc cc}
\toprule
\multirow{3}{*}{\textbf{Model}} & \multicolumn{6}{c}{\textbf{Vehicle}} & \multicolumn{6}{c}{\textbf{Pedestrian}} \\
\cmidrule(lr){2-7} \cmidrule(lr){8-13}
 & \multicolumn{2}{c}{\textbf{Close}} & \multicolumn{2}{c}{\textbf{Suitable}} & \multicolumn{2}{c}{\textbf{Far}} & \multicolumn{2}{c}{\textbf{Close}} & \multicolumn{2}{c}{\textbf{Suitable}} & \multicolumn{2}{c}{\textbf{Far}} \\
\cmidrule(lr){2-3} \cmidrule(lr){4-5} \cmidrule(lr){6-7} \cmidrule(lr){8-9} \cmidrule(lr){10-11} \cmidrule(lr){12-13}
 & \textbf{ATF} & \textbf{SR (\%)} & \textbf{ATF} & \textbf{SR (\%)} & \textbf{ATF} & \textbf{SR (\%)} & \textbf{ATF} & \textbf{SR (\%)} & \textbf{ATF} & \textbf{SR (\%)} & \textbf{ATF} & \textbf{SR (\%)} \\
\midrule
ACT & 34.62 & 0 & 32.33 & 0 & 34.15 & 0 & 19.30 & 0 & 19.20 & 0 & 18.25 & 0 \\
WALL-OSS & 23.33 & 0 & 44.59 & 0 & 28.54 & 0 & 50.14 & 0 & 52.86 & 0 & 31.60 & 0 \\
$\pi_0$ & 75.50 & 10 & 98.50 & 11.11 & 91.60 & 0 & 70.67 & 0 & 108.10 & 20 & 132.33 & 33.33 \\
w/o Aux. Head & 106.88 & 12.50 & 127.20 & 13.33 & 123.67 & 8.33 & \textbf{127.40} & \textbf{20} & 130 & \textbf{25} & 157.14 & 28.57 \\
$\pi_{0.5}$ & 80.62 & 0 & \textbf{178.33} & \textbf{33.33} & 177.30 & \textbf{30} & 101.11 & 11.11 & 195.83 & 33.33 & 193.57 & 42.86 \\
\midrule
\textbf{UAV-Track VLA} & \textbf{111.25} & \textbf{16.67} & 125.52 & 12.12 & \textbf{185} & 26.67 & 182.12 & \textbf{25} & \textbf{200.47} & 41.18 & \textbf{262.20} & \textbf{60} \\
\bottomrule
\end{tabular}
}
\end{table}

\begin{table}[htbp]
\centering
\caption{Detailed Tracking Performance on \textbf{Town05} (Seen Map).}
\label{tab:perf_town05}
\resizebox{\textwidth}{!}{
\begin{tabular}{l cc cc cc cc cc cc}
\toprule
\multirow{3}{*}{\textbf{Model}} & \multicolumn{6}{c}{\textbf{Vehicle}} & \multicolumn{6}{c}{\textbf{Pedestrian}} \\
\cmidrule(lr){2-7} \cmidrule(lr){8-13}
 & \multicolumn{2}{c}{\textbf{Close}} & \multicolumn{2}{c}{\textbf{Suitable}} & \multicolumn{2}{c}{\textbf{Far}} & \multicolumn{2}{c}{\textbf{Close}} & \multicolumn{2}{c}{\textbf{Suitable}} & \multicolumn{2}{c}{\textbf{Far}} \\
\cmidrule(lr){2-3} \cmidrule(lr){4-5} \cmidrule(lr){6-7} \cmidrule(lr){8-9} \cmidrule(lr){10-11} \cmidrule(lr){12-13}
 & \textbf{ATF} & \textbf{SR (\%)} & \textbf{ATF} & \textbf{SR (\%)} & \textbf{ATF} & \textbf{SR (\%)} & \textbf{ATF} & \textbf{SR (\%)} & \textbf{ATF} & \textbf{SR (\%)} & \textbf{ATF} & \textbf{SR (\%)} \\
\midrule
ACT & 54.31 & 0 & 50.31 & 0 & 47.38 & 0 & 18.29 & 0 & 18.29 & 0 & 18.71 & 0 \\
WALL-OSS & 27 & 0 & 44.15 & 0 & 45.69 & 0 & 36.86 & 0 & 57.14 & 0 & 40.71 & 0 \\
$\pi_0$ & 64.56 & 0 & 116.50 & 8.33 & 95.92 & 0 & 92.17 & 0 & 167.71 & 28.57 & 146.80 & 20 \\
w/o Aux. Head & 110.15 & 7.69 & 221.77 & 38.46 & 231.05 & \textbf{75} & 109 & 0 & 186.17 & 0 & 240 & 40 \\
$\pi_{0.5}$ & 116.85 & 15.38 & 208.38 & 46.15 & \textbf{280.50} & 66.67 & 100.83 & 0 & 222.83 & 33.33 & 276.30 & 60 \\
\midrule
\textbf{UAV-Track VLA} & \textbf{126.33} & \textbf{16.67} & \textbf{234.77} & \textbf{48.57} & 187.27 & 36.36 & \textbf{156} & \textbf{14.29} & \textbf{329.40} & \textbf{80} & \textbf{284} & \textbf{77.78} \\
\bottomrule
\end{tabular}
}
\end{table}

\begin{table}[htbp]
\centering
\caption{Detailed Tracking Performance on \textbf{Town06} (Seen Map).}
\label{tab:perf_town06}
\resizebox{\textwidth}{!}{
\begin{tabular}{l cc cc cc cc cc cc}
\toprule
\multirow{3}{*}{\textbf{Model}} & \multicolumn{6}{c}{\textbf{Vehicle}} & \multicolumn{6}{c}{\textbf{Pedestrian}} \\
\cmidrule(lr){2-7} \cmidrule(lr){8-13}
 & \multicolumn{2}{c}{\textbf{Close}} & \multicolumn{2}{c}{\textbf{Suitable}} & \multicolumn{2}{c}{\textbf{Far}} & \multicolumn{2}{c}{\textbf{Close}} & \multicolumn{2}{c}{\textbf{Suitable}} & \multicolumn{2}{c}{\textbf{Far}} \\
\cmidrule(lr){2-3} \cmidrule(lr){4-5} \cmidrule(lr){6-7} \cmidrule(lr){8-9} \cmidrule(lr){10-11} \cmidrule(lr){12-13}
 & \textbf{ATF} & \textbf{SR (\%)} & \textbf{ATF} & \textbf{SR (\%)} & \textbf{ATF} & \textbf{SR (\%)} & \textbf{ATF} & \textbf{SR (\%)} & \textbf{ATF} & \textbf{SR (\%)} & \textbf{ATF} & \textbf{SR (\%)} \\
\midrule
ACT & 34.29 & 0 & 28.53 & 0 & 33 & 0 & 19.80 & 0 & 24.32 & 0 & 24.40 & 0 \\
WALL-OSS & 25.14 & 0 & 42.26 & 0 & 35.85 & 0 & 43 & 0 & 62.48 & 0 & 43.16 & 0 \\
$\pi_0$ & \textbf{67.64} & 0 & 66.69 & 0 & 94.80 & 10 & 107.17 & 0 & 222.33 & \textbf{33.33} & 137 & 21.43 \\
w/o Aux. Head & 65.50 & \textbf{16.67} & 129.50 & 20 & 120.30 & 0 & 180.40 & 10 & 187.36 & 28 & 168.16 & 32 \\
$\pi_{0.5}$ & 46.45 & 0 & 96.29 & 14.29 & 90.17 & 8.33 & 101 & 0 & 189.09 & 27.27 & 175.76 & 36.36 \\
\midrule
\textbf{UAV-Track VLA} & 52 & 0 & \textbf{171.16} & \textbf{32} & \textbf{251.20} & \textbf{66.67} & \textbf{286.90} & \textbf{60} & \textbf{314.16} & 76 & \textbf{383.60} & \textbf{80} \\
\bottomrule
\end{tabular}
}
\end{table}

\begin{table}[htbp]
\centering
\caption{Detailed Tracking Performance on \textbf{Town07} (Seen Map).}
\label{tab:perf_town07}
\resizebox{\textwidth}{!}{
\begin{tabular}{l cc cc cc cc cc cc}
\toprule
\multirow{3}{*}{\textbf{Model}} & \multicolumn{6}{c}{\textbf{Vehicle}} & \multicolumn{6}{c}{\textbf{Pedestrian}} \\
\cmidrule(lr){2-7} \cmidrule(lr){8-13}
 & \multicolumn{2}{c}{\textbf{Close}} & \multicolumn{2}{c}{\textbf{Suitable}} & \multicolumn{2}{c}{\textbf{Far}} & \multicolumn{2}{c}{\textbf{Close}} & \multicolumn{2}{c}{\textbf{Suitable}} & \multicolumn{2}{c}{\textbf{Far}} \\
\cmidrule(lr){2-3} \cmidrule(lr){4-5} \cmidrule(lr){6-7} \cmidrule(lr){8-9} \cmidrule(lr){10-11} \cmidrule(lr){12-13}
 & \textbf{ATF} & \textbf{SR (\%)} & \textbf{ATF} & \textbf{SR (\%)} & \textbf{ATF} & \textbf{SR (\%)} & \textbf{ATF} & \textbf{SR (\%)} & \textbf{ATF} & \textbf{SR (\%)} & \textbf{ATF} & \textbf{SR (\%)} \\
\midrule
ACT & 36.09 & 0 & 32.55 & 0 & 34.64 & 0 & 18.50 & 0 & 20.89 & 0 & 18.58 & 0 \\
WALL-OSS & 23 & 0 & 44.10 & 0 & 32.73 & 0 & 50.17 & 0 & 56.67 & 0 & 68.68 & 0 \\
$\pi_0$ & 71.87 & 0 & 119.50 & 16.67 & 102.50 & 6.25 & 83 & 0 & 149 & 16.67 & 140.80 & 20 \\
w/o Aux. Head & 65.25 & 0 & 124.62 & 15.38 & 128.92 & 30.77 & 172.67 & 16.67 & 182.26 & 26.32 & 196.21 & 36.84 \\
$\pi_{0.5}$ & 81.16 & 10.53 & 120.60 & 6.67 & 119.83 & 16.67 & \textbf{293} & \textbf{50} & 243.50 & 50 & 249.05 & 50 \\
\midrule
\textbf{UAV-Track VLA} & \textbf{110.55} & \textbf{18.18} & \textbf{131.55} & \textbf{19.35} & \textbf{147.67} & \textbf{16.67} & 179 & 33.33 & \textbf{291.84} & \textbf{68.42} & \textbf{278.38} & \textbf{50} \\
\bottomrule
\end{tabular}
}
\end{table}

\begin{table}[htbp]
\centering
\caption{Detailed Tracking Performance on \textbf{Town10} (Seen Map).}
\label{tab:perf_town10}
\resizebox{\textwidth}{!}{
\begin{tabular}{l cc cc cc cc cc cc}
\toprule
\multirow{3}{*}{\textbf{Model}} & \multicolumn{6}{c}{\textbf{Vehicle}} & \multicolumn{6}{c}{\textbf{Pedestrian}} \\
\cmidrule(lr){2-7} \cmidrule(lr){8-13}
 & \multicolumn{2}{c}{\textbf{Close}} & \multicolumn{2}{c}{\textbf{Suitable}} & \multicolumn{2}{c}{\textbf{Far}} & \multicolumn{2}{c}{\textbf{Close}} & \multicolumn{2}{c}{\textbf{Suitable}} & \multicolumn{2}{c}{\textbf{Far}} \\
\cmidrule(lr){2-3} \cmidrule(lr){4-5} \cmidrule(lr){6-7} \cmidrule(lr){8-9} \cmidrule(lr){10-11} \cmidrule(lr){12-13}
 & \textbf{ATF} & \textbf{SR (\%)} & \textbf{ATF} & \textbf{SR (\%)} & \textbf{ATF} & \textbf{SR (\%)} & \textbf{ATF} & \textbf{SR (\%)} & \textbf{ATF} & \textbf{SR (\%)} & \textbf{ATF} & \textbf{SR (\%)} \\
\midrule
ACT & 23.57 & 0 & 20.29 & 0 & 14.50 & 0 & 14 & 0 & 19 & 0 & 17 & 0 \\
WALL-OSS & 23.43 & 0 & 73 & 0 & 23 & 0 & 36 & 0 & 50.12 & 0 & 28 & 0 \\
$\pi_0$ & 60.06 & 0 & 75.60 & 0 & 54.91 & 0 & 86.20 & 0 & 125.50 & 0 & 112.43 & 0 \\
w/o Aux. Head & 61.85 & 7.69 & 147.60 & 26.67 & 158.17 & \textbf{41.67} & 108.50 & 0 & 168.62 & 25 & 186.75 & 37.50 \\
$\pi_{0.5}$ & 75 & 0 & 107.71 & 17.65 & 136.89 & 22.22 & 75.20 & 0 & 144.12 & 12.50 & \textbf{218.38} & \textbf{50} \\
\midrule
\textbf{UAV-Track VLA} & \textbf{82.90} & \textbf{10} & \textbf{154.71} & \textbf{35.71} & \textbf{191.46} & 38.46 & \textbf{194.40} & \textbf{20} & \textbf{185.73} & \textbf{36.36} & 165.14 & 42.86 \\
\bottomrule
\end{tabular}
}
\end{table}


\begin{table}[htbp]
\centering
\caption{Detailed Tracking Performance on \textbf{Town01} (Unseen Map).}
\label{tab:perf_town01}
\resizebox{\textwidth}{!}{
\begin{tabular}{l cc cc cc cc cc cc}
\toprule
\multirow{3}{*}{\textbf{Model}} & \multicolumn{6}{c}{\textbf{Vehicle}} & \multicolumn{6}{c}{\textbf{Pedestrian}} \\
\cmidrule(lr){2-7} \cmidrule(lr){8-13}
 & \multicolumn{2}{c}{\textbf{Close}} & \multicolumn{2}{c}{\textbf{Suitable}} & \multicolumn{2}{c}{\textbf{Far}} & \multicolumn{2}{c}{\textbf{Close}} & \multicolumn{2}{c}{\textbf{Suitable}} & \multicolumn{2}{c}{\textbf{Far}} \\
\cmidrule(lr){2-3} \cmidrule(lr){4-5} \cmidrule(lr){6-7} \cmidrule(lr){8-9} \cmidrule(lr){10-11} \cmidrule(lr){12-13}
 & \textbf{ATF} & \textbf{SR (\%)} & \textbf{ATF} & \textbf{SR (\%)} & \textbf{ATF} & \textbf{SR (\%)} & \textbf{ATF} & \textbf{SR (\%)} & \textbf{ATF} & \textbf{SR (\%)} & \textbf{ATF} & \textbf{SR (\%)} \\
\midrule
ACT & 22.80 & 0 & 28.88 & 0 & 34.64 & 0 & 17.44 & 0 & 16.57 & 0 & 17.71 & 0 \\
WALL-OSS & 27.77 & 0 & 48 & 0 & 29.45 & 0 & 46.62 & 0 & 41.33 & 0 & 43 & 0 \\
$\pi_0$ & \textbf{85.60} & 0 & 72.17 & 0 & 97.53 & 6.67 & 99.67 & 0 & 130.44 & 11.11 & 118.40 & 20 \\
w/o Aux. Head & 74.25 & 0 & 148.07 & 14.29 & 132.50 & 12.50 & 92.33 & 11.11 & 139 & 12.50 & 148.20 & 20 \\
$\pi_{0.5}$ & 81.21 & 0 & \textbf{244.73} & \textbf{36.36} & 148.75 & 33.33 & 134.75 & 25 & 83.88 & 0 & 113.71 & 0 \\
\midrule
\textbf{UAV-Track VLA} & 75.15 & \textbf{7.69} & 133.96 & 22.22 & \textbf{181.85} & \textbf{38.46} & \textbf{150.14} & \textbf{28.57} & \textbf{199.30} & \textbf{30.43} & \textbf{169.71} & \textbf{42.86} \\
\bottomrule
\end{tabular}
}
\end{table}

\begin{table}[htbp]
\centering
\caption{Detailed Tracking Performance on \textbf{Town03} (Unseen Map).}
\label{tab:perf_town03}
\resizebox{\textwidth}{!}{
\begin{tabular}{l cc cc cc cc cc cc}
\toprule
\multirow{3}{*}{\textbf{Model}} & \multicolumn{6}{c}{\textbf{Vehicle}} & \multicolumn{6}{c}{\textbf{Pedestrian}} \\
\cmidrule(lr){2-7} \cmidrule(lr){8-13}
 & \multicolumn{2}{c}{\textbf{Close}} & \multicolumn{2}{c}{\textbf{Suitable}} & \multicolumn{2}{c}{\textbf{Far}} & \multicolumn{2}{c}{\textbf{Close}} & \multicolumn{2}{c}{\textbf{Suitable}} & \multicolumn{2}{c}{\textbf{Far}} \\
\cmidrule(lr){2-3} \cmidrule(lr){4-5} \cmidrule(lr){6-7} \cmidrule(lr){8-9} \cmidrule(lr){10-11} \cmidrule(lr){12-13}
 & \textbf{ATF} & \textbf{SR (\%)} & \textbf{ATF} & \textbf{SR (\%)} & \textbf{ATF} & \textbf{SR (\%)} & \textbf{ATF} & \textbf{SR (\%)} & \textbf{ATF} & \textbf{SR (\%)} & \textbf{ATF} & \textbf{SR (\%)} \\
\midrule
ACT & 38 & 0 & 43.83 & 0 & 43.25 & 0 & 19.50 & 0 & 21 & 0 & 22.29 & 0 \\
WALL-OSS & 21.78 & 0 & 57.29 & 0 & 33.53 & 0 & 32.75 & 0 & 45.33 & 0 & 36.43 & 0 \\
$\pi_0$ & 68.67 & 0 & 92.57 & 14.29 & 60.53 & 0 & 85.29 & 14.29 & 136.67 & 0 & 155.67 & 33.33 \\
w/o Aux. Head & 61.82 & 9.09 & \textbf{207.43} & \textbf{50} & 105.69 & 7.69 & 86.89 & 11.11 & 125.86 & 0 & 241.17 & 50 \\
$\pi_{0.5}$ & \textbf{97.47} & \textbf{6.67} & 197.53 & 35.29 & \textbf{136.09} & \textbf{18.18} & 81 & 0 & 175 & \textbf{40} & 159.86 & 14.29 \\
\midrule
\textbf{UAV-Track VLA} & 72.53 & 6.67 & 165.33 & 33.33 & 151.18 & 23.53 & \textbf{150.33} & \textbf{33.33} & \textbf{187.67} & 33.33 & \textbf{295} & \textbf{71.43} \\
\bottomrule
\end{tabular}
}
\end{table}

\begin{table}[htbp]
\centering
\caption{Detailed Tracking Performance on \textbf{Town04} (Unseen Map).}
\label{tab:perf_town04}
\resizebox{\textwidth}{!}{
\begin{tabular}{l cc cc cc cc cc cc}
\toprule
\multirow{3}{*}{\textbf{Model}} & \multicolumn{6}{c}{\textbf{Vehicle}} & \multicolumn{6}{c}{\textbf{Pedestrian}} \\
\cmidrule(lr){2-7} \cmidrule(lr){8-13}
 & \multicolumn{2}{c}{\textbf{Close}} & \multicolumn{2}{c}{\textbf{Suitable}} & \multicolumn{2}{c}{\textbf{Far}} & \multicolumn{2}{c}{\textbf{Close}} & \multicolumn{2}{c}{\textbf{Suitable}} & \multicolumn{2}{c}{\textbf{Far}} \\
\cmidrule(lr){2-3} \cmidrule(lr){4-5} \cmidrule(lr){6-7} \cmidrule(lr){8-9} \cmidrule(lr){10-11} \cmidrule(lr){12-13}
 & \textbf{ATF} & \textbf{SR (\%)} & \textbf{ATF} & \textbf{SR (\%)} & \textbf{ATF} & \textbf{SR (\%)} & \textbf{ATF} & \textbf{SR (\%)} & \textbf{ATF} & \textbf{SR (\%)} & \textbf{ATF} & \textbf{SR (\%)} \\
\midrule
ACT & 33.25 & 0 & 28.33 & 0 & 39.15 & 0 & 19.25 & 0 & 18.75 & 0 & 19.14 & 0 \\
WALL-OSS & 36.92 & 0 & 32.58 & 0 & 39.93 & 0 & 42.14 & 0 & 65.29 & 0 & 46.33 & 0 \\
$\pi_0$ & 75.62 & 6.25 & 117.64 & 14.29 & 72.67 & 0 & 65.29 & 0 & 213 & \textbf{66.67} & 157.80 & 0 \\
w/o Aux. Head & 74.85 & 0 & 166.69 & 15.38 & 122.08 & \textbf{30.77} & 119 & 16.67 & 146.12 & 12.50 & 96.86 & 0 \\
$\pi_{0.5}$ & 68.67 & \textbf{8.33} & 165.18 & 27.27 & 109.67 & 5.56 & \textbf{158.12} & \textbf{37.50} & 162.12 & 37.50 & 150.33 & 0 \\
\midrule
\textbf{UAV-Track VLA} & \textbf{78.53} & 5.88 & \textbf{176.82} & \textbf{45.45} & \textbf{149.15} & 23.08 & 147.14 & 28.57 & \textbf{286} & \textbf{66.67} & \textbf{214.17} & \textbf{50} \\
\bottomrule
\end{tabular}
}
\end{table}

\end{document}